\documentclass[10pt,twocolumn,letterpaper]{article}

\usepackage{cvpr}
\usepackage{graphicx}
\usepackage{amsmath}
\usepackage{amssymb}
\usepackage{booktabs}
\usepackage{multirow}
\usepackage[pagebackref,breaklinks,colorlinks]{hyperref}
\usepackage[capitalize]{cleveref}
\crefname{section}{Sec.}{Secs.}
\Crefname{section}{Section}{Sections}
\Crefname{table}{Table}{Tables}
\crefname{table}{Tab.}{Tabs.}

\usepackage{algorithm}
\usepackage{algpseudocode}
\usepackage{newfloat}
\usepackage{listings}

\def\confName{CVPR}
\def\confYear{2025}

\title{
Static Scene Reconstruction 
from Dynamic Egocentric Videos
}

\author{
Qifei Cui \hspace{5pt} Patrick Chen\\
Department of Computer and Information Science\\  University of Pennsylvania
}

\begin{document}
\maketitle
\begin{abstract}
Egocentric videos present unique challenges for 3D reconstruction due to rapid camera motion and frequent dynamic interactions. State-of-the-art static reconstruction systems, such as MapAnything, often degrade in these settings, suffering from catastrophic trajectory drift and "ghost" geometry caused by moving hands. We bridge this gap by proposing a robust pipeline that adapts static reconstruction backbones to long-form egocentric video. Our approach introduces a mask-aware reconstruction mechanism that explicitly suppresses dynamic foreground in the attention layers, preventing hand artifacts from contaminating the static map. Furthermore, we employ a chunked reconstruction strategy with pose-graph stitching to ensure global consistency and eliminate long-term drift. Experiments on HD-EPIC and indoor drone datasets demonstrate that our pipeline significantly improves absolute trajectory error and yields visually clean static geometry compared to naive baselines, effectively extending the capability of foundation models to dynamic first-person scenes.

\end{abstract}    
\section{Introduction}
\label{sec:intro}

Egocentric video offers a unique, richly structured view of human activities: hands manipulate objects at close range, tools occlude and reveal surfaces, and the camera undergoes aggressive motion through cluttered environments.
These properties make egocentric streams an attractive signal for learning object-centric representations, affordances, and downstream manipulation policies.
However, they also pose severe challenges for 3D reconstruction.

Recent methods such as MapAnything~\cite{keetha2025mapanything} demonstrate impressive performance on static scenes, providing dense metric geometry and accurate camera trajectories.
Yet, when applied to dynamic, first-person videos from datasets such as HD-EPIC~\cite{HDEPIC} or indoor drone footage, we observe systematic failures:
(i) long sequences accumulate drift, flattening entire buildings;
(ii) camera initialization is fragile, sometimes incurring $180^\circ$ orientation flips;
and (iii) large, frequently moving hands and objects are interpreted as part of the static world, leading to severe ghosting and physically impossible geometry.

In this paper, we ask:

\begin{quote}
\emph{How can we adapt a static reconstruction backbone like MapAnything to egocentric video, such that we obtain a reliable static map while explicitly accounting for dynamic foreground without retraining or funing the model?}
\end{quote}

Our answer is a mask-aware, chunked pipeline that progressively stabilizes static geometry and isolates dynamic content. We build our story around the concrete experience of deploying MapAnything on the most recent representative egocentric settings HD-EPIC kitchen activities with strong hand motion.

\begin{figure*}[t]
    \centering
    \includegraphics[width=\textwidth]{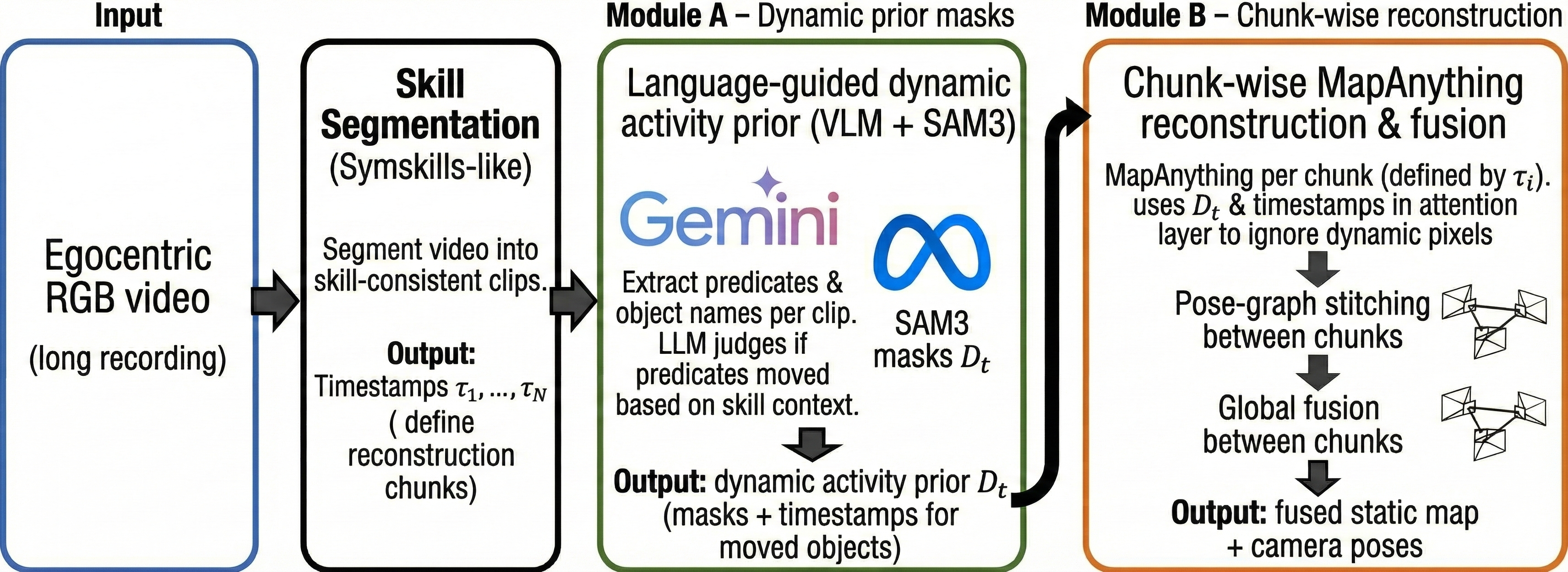}
    \caption{\textbf{Overall pipeline.} \textbf{Module A} constructs a binary
    dynamic activity prior $D_t$ from VLM (Gemini) prompts and SAM3 video
    segmentation/tracking, optionally gated by interaction time windows.
    \textbf{Module B} runs chunk-wise reconstruction and fusion (MapAnything)
    while ignoring dynamic pixels indicated by $D_t$, then stitches chunks using
    overlap-based Sim(3) alignment to obtain globally consistent poses and a
    fused static map.}
    \label{fig:overall-pipeline}
\end{figure*}
\section{Background Information}
\label{sec:background}
\textbf{EPIC-KITCHENS HD (HD-EPIC)}
HD-EPIC is a large-scale egocentric video dataset comprising 41 hours of unscripted cooking activities captured from a first-person perspective. The dataset features rich semantic annotations, including 59,000 fine-grained action segments and over 20,000 object interaction events accompanied by 3D masks and trajectories \cite{HDEPIC}. Furthermore, HD-EPIC provides 3D reconstructions of kitchen environments, enabling spatially grounded analysis over time.

\textbf{Static scene reconstruction.} 
Transformer-based models like \textit{MapAnything} have recently shown impressive results in regressing 3D geometry directly from egocentric video frames. By predicting dense depth maps and camera poses in a single forward pass, MapAnything produces metrically accurate spatial reconstructions \cite{keetha2025mapanything}. These capabilities enable spatially grounded understanding of object-object relations, crucial for manipulation tasks. However, the method assumes static environments and often fails in highly dynamic first-person scenes, where foreground motion—particularly hands—causes ghosting artifacts or catastrophic pose drift. This fragility is especially evident in long sequences from datasets like HD-EPIC \cite{HDEPIC}, where moving hands are incorrectly fused into the static map. Therefore, to relief these, we need a temporal consistent object tracker.

\textbf{Temporal object tracking with SAM3.} 
To isolate dynamic content, we employ \textit{Segment Anything Model 3 (SAM3)}, an open-vocabulary segmentation model trained on over 4 million concept-labeled masks \cite{carion2025sam3segmentconcepts}. SAM3 extends Meta AI’s original SAM with DETR-style detection and a memory-based transformer tracker, enabling prompt-driven segmentation and long-term instance tracking. Its support for both text and image prompts allows flexible identification of dynamic agents—such as hands and tools—in egocentric videos. Beyond generating temporally consistent masks, SAM3 can also models object appearance, making it well-suited for mask-aware dynamic filtering.

\textbf{Egocentric data as a learning signal.} 
Egocentric video data offers distinct advantages over third-person perspectives by focusing on the wearer's field of interaction—primarily hands and objects—thereby capturing subtle human-object relations \cite{deng2025egocentrichumanobjectinteractiondetection}. Prior research demonstrates that modeling the hand's shape and viewpoint in 2D egocentric frames significantly enhances object recognition performance \cite{ObjectRecognitionLeveragingthe3DShapeGraspingHand}. These insights motivate the extension to 3D egocentric modeling, where methods like MapAnything can integrate metric spatial context with SAM3’s concept-level segmentation. In our project, we aim to fuse static scene geometry from MapAnything with temporally consistent dynamic object tracking via SAM3. This integration allows us to mask out dynamic manipulators such as hands and tools during 3D reconstruction, mitigating artifacts and pose drift in highly interactive settings—ultimately enabling robust reconstruction pipelines for real-world egocentric video.

\section{Methodology}
\label{sec:method}

Our goal is to reconstruct a \emph{static} scene map, which has a fused point cloud with a globally aligned camera trajectory, from long egocentric RGB videos that contain frequent hands and manipulated objects. A key failure mode of standard monocular reconstruction on such data is that dynamic foreground regions still produce visually consistent features across frames (hands, grasped objects, reflections, etc.), leading the reconstructor to form incorrect cross-frame correspondences and to aggregate these regions as if they were static. This results in characteristic ``ghosting'' artifacts and, more critically for chunked pipelines, unstable pose/scale estimates that accumulate into cross-chunk alignment drift.

We mitigate this by injecting an explicit \emph{interaction-aware} dynamic prior into the reconstruction pipeline. To build a global static map, we use (A) per-frame instance masks from SAM3 initialized by language cues to identify candidate dynamic entities, (B) precise interaction onset times from a skill/interaction segmentation pipeline to \emph{activate} masks only after motion begins and keep them active thereafter, and (C) chunk-wise reconstruction with overlap-based Sim(3) stitching, where we estimate the similarity transform between consecutive chunks via Umeyama alignment on overlapping camera centers.

\subsection{Problem Setup}
\label{sec:method-problem}

Let $\{I_t\}_{t=1}^{T}$ be a long egocentric RGB video. Our goal is to recover a \emph{static} 3D scene representation (a fused dense point cloud) from this sequence, despite frequent hands and manipulated objects. We do \emph{not} assume access to globally consistent poses a priori. In principle one could process the entire video jointly. However, MapAnything's video reconstructor operates on dense 2D tokens with temporal attention, making end-to-end inference on long sequences prohibitively expensive in memory and runtime (and often infeasible on commodity GPUs). We therefore process the video in overlapping temporal chunks, obtain chunk-local reconstructions, and construct a global frame via overlap-based alignment.

\paragraph{Interaction-aware dynamic prior.}
We construct a per-frame binary suppression mask $D_t\in\{0,1\}^{H\times W}$ that combines two components: (i) a \emph{hand} mask that is active throughout the sequence, and (ii) an \emph{object} mask that becomes active only after an interaction onset and then remains active cumulatively. Specifically, a
SymSkill-style interaction/skill segmentor partitions the long video into short clips and outputs subject-predicate-object (S-P-O) predictions. An LLM then selects motion-inducing predicates to identify candidate objects and their interaction onset times. Given these candidates, a VLM provides coarse
localization cues (tags/boxes) that are converted into prompt specifications and fed to SAM3, which tracks instances over time and returns per-frame masks with persistent IDs. Using the onset time $a_i$ for each tracked object instance, we activate object masks only when $t\ge a_i$ (cumulative thereafter) and finally union them with the always-on hand mask to form $D_t$. The pixel-level mask $D_t$ is further mapped to the tokenizer's 2D token grid to produce a token indicator used for attention masking in MapAnything.

\paragraph{Chunked reconstruction (with dynamic priors).}
We partition the video into overlapping chunks of length $K$ with overlap $O$ frames (in our experiments $K{=}192,\,O{=}96$), and run MapAnything per chunk while injecting the token-level dynamic prior (as an attention mask over 2D tokens) to suppress dynamic regions during inference. For each chunk $c$, MapAnything outputs per-frame depth and camera intrinsics (estimated internally), together with chunk-local camera poses $\{\hat{T}^{(c)}_t\}$ and a dense point cloud $\hat{\mathcal{P}}^{(c)}$ expressed
in the chunk coordinate frame.

\paragraph{Global frame construction via Umeyama stitching.}
Because each chunk reconstruction is defined up to an unknown similarity transform, we align consecutive chunks using the overlapping frames. Let $S_c\in \mathrm{Sim}(3)$ denote the similarity transform that maps chunk $c$ into the global frame. We estimate $S_c$ with Umeyama alignment on the overlap camera centers/trajectory, then obtain global poses and a fused static map by $T_t := S_{c(t)}\hat{T}^{(c(t))}_t$ and $\mathcal{P}:=\bigcup_c S_c(\hat{\mathcal{P}}^{(c)})$, where the global frame is initialized by setting $S_{c=1}$ as the identity (the first chunk as the origin).

\subsection{Module A: Dynamic Prior Masks}
\label{sec:method-dynamic-prior}

\begin{figure*}[t]
  \centering
  \includegraphics[width=\linewidth]{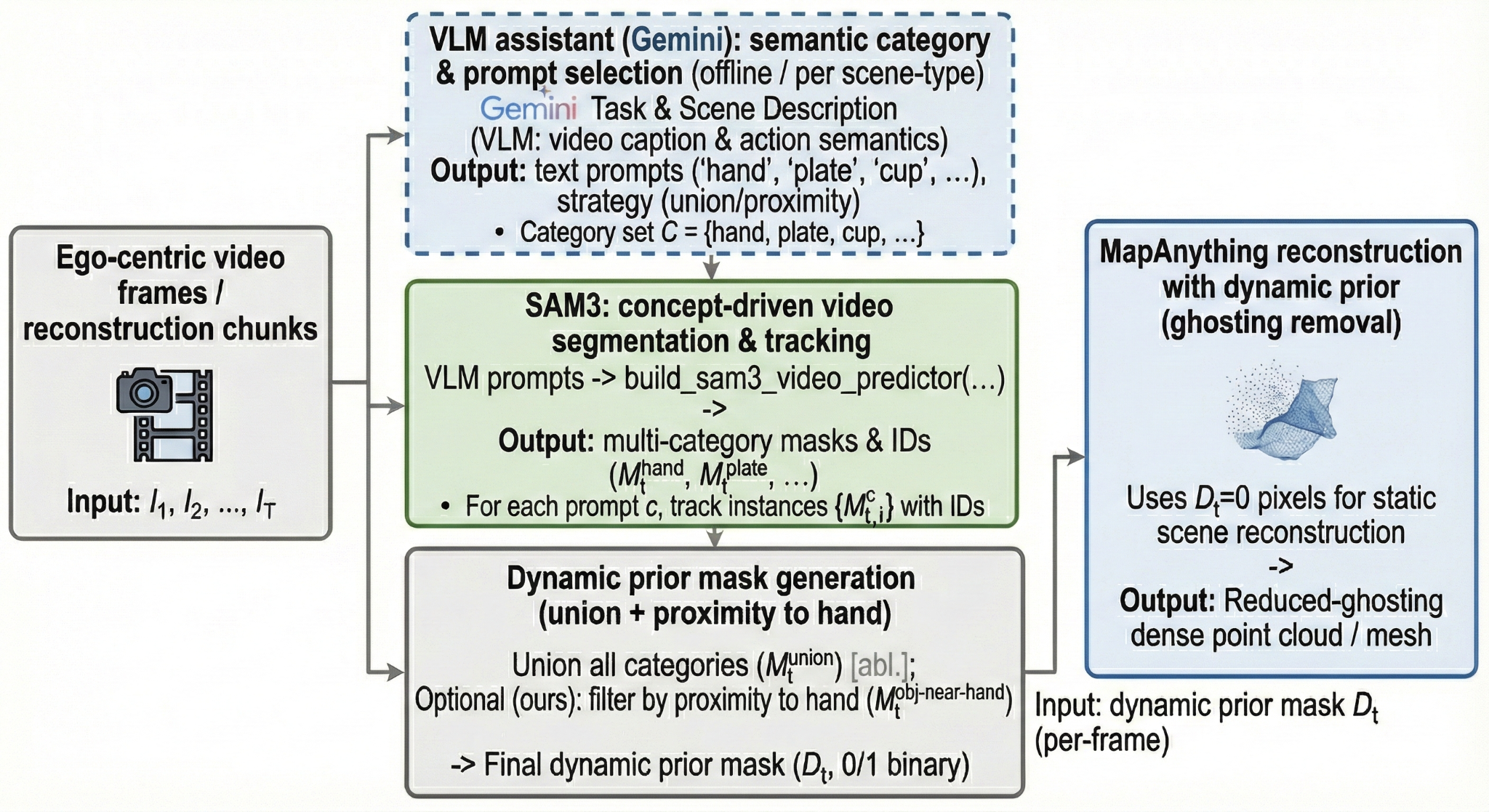}
  \caption{\textbf{Module A: interaction-aware dynamic prior pipeline.}
  Given egocentric video frames (or reconstruction chunks), a VLM selects scene-relevant semantic categories and text prompts (e.g., \emph{hand}, \emph{plate}, \emph{cup}). 
  SAM3 then performs concept-driven video segmentation and tracking to produce per-category instance masks with consistent IDs.
  We aggregate these masks into a per-frame dynamic prior $D_t$ (union over categories, optionally filtered by proximity to the hand to isolate manipulated objects).
  Finally, MapAnything consumes $D_t$ to suppress dynamic regions and reconstruct the static scene with reduced ghosting artifacts.}
  \label{fig:pipeline_moduleA}
\end{figure*}

Module~A constructs an \emph{interaction-aware} dynamic prior that identifies (i) hands to be suppressed throughout the sequence and (ii) manipulated objects to be suppressed only after they begin to move. The resulting per-frame binary mask is later mapped to the 2D token grid and injected as an attention mask in MapAnything (\S\ref{sec:method-recon}). An overview of the complete pipeline in Module~A is illustrated in Fig.~\ref{fig:pipeline_moduleA}.

\paragraph{A1: Skill-segmented clips and motion-inducing object candidates.}
We first segment the long video into short interaction-centric clips using a SymSkill skill segmentor. For each clip, the segmentor produces
subject-predicate-object (S-P-O) predictions. We then use an LLM to select \emph{motion-inducing} predicates (e.g., \texttt{pick\_up}, \texttt{place}, \texttt{open}, \texttt{close}, \texttt{pour}) and extract the corresponding object candidates that are likely to undergo pose change. This step yields a set of candidate object names per clip and an interaction onset time for each candidate instance.

\paragraph{A2: VLM localization and a unified prompt schema.}
Given each clip and its candidate object list, we localize these objects via a VLM that outputs coarse spatial priors. We represent prompts in a unified JSON schema with fields \texttt{\{track\_id, name, box=(x,y,w,h)\}} and optionally sparse point prompts \texttt{\{(x,y)\}} when only point cues are available. These prompts are used to initialize instance-conditioned tracking in SAM3. When needed, we provide lightweight human-in-the-loop sparse point prompts on a few keyframes to initialize or correct tracks under occlusion and small-object drift.

\paragraph{A3: SAM3 video segmentation and persistent instance tracking.}
Conditioned on the prompts from Step~A2 (VLM-derived coarse localization), and optionally refined via human-in-the-loop sparse clicks, SAM3 performs video segmentation and returns per-frame instance masks with \emph{persistent} IDs:
\begin{equation}
    \mathcal{M}_t \;=\; \bigl\{ M_{t,i} \in \{0,1\}^{H\times W} \bigr\}_{i=1}^{N_t},
\end{equation}
where $i$ indexes tracked instances (objects and hands). In practice, prompts may be provided on multiple keyframes via lightweight human-in-the-loop point clicks (e.g., to initialize small objects, handle brief occlusions, or correct drift). We inject \emph{all} human-refined keyframes into the SAM3 inference session before running video propagation, so that later key frame refinements are respected when producing the final mask tracks. The output of this step is thus a temporally aligned set of instance masks $\{M_{t,i}\}$ suitable for downstream interaction activation and binary prior construction.

\paragraph{A4: Interaction onset times and cumulative activation.}
Let $a_i \in \{1,\dots,T\}$ denote the interaction onset time of object instance $i$ (from the skill segmentor; EPIC-HD interaction annotations can be used as an oracle for rapid iteration). We use \emph{cumulative} activation: once an object is first touched or moved, we treat it as dynamic and mask it for the remainder of the video. Define the activated instance set
\begin{equation}
    \mathcal{A}_t=\{i \mid a_i \le t\}.
\end{equation}
Cumulative masking prioritizes static-map purity by preventing a moved object from re-entering the static map after its pose changes even when objects return to their original pose.

\paragraph{A5: From instance masks to a binary dynamic prior.}
We form two components: a \emph{hand} mask that is always active and an \emph{object} mask that is activated only after onset. Let $D^{\text{hand}}_t$ be the union of all hand instance masks at time $t$,
\begin{equation}
    D^{\text{hand}}_t(\mathbf{x}) \;=\;
    \mathbf{1}\Bigl\{ \mathbf{x} \in \bigcup_{i\in \mathcal{H}_t} M_{t,i} \Bigr\},
\end{equation}
where $\mathcal{H}_t$ denotes the set of hand instances at time $t$. For manipulated objects, the union baseline aggregates only \emph{activated} instances:
\begin{equation}
D^{\text{obj}}_t(\mathbf{x}) \;=\;
\mathbf{1}\Bigl\{ \mathbf{x} \in \bigcup_{i \in \mathcal{A}_t} M_{t,i} \Bigr\}.
\label{eq:obj-union}
\end{equation}

Optionally, we restrict activated object instances using a hand-centric proximity filter (an ablation that reduces false positives when multiple similar objects are present). Let $M^{\text{hand}}_{t}$ be the union of all hand masks at time $t$. We dilate the hand mask by $r$ pixels and keep instance $i$ if it overlaps the dilated hand region above a threshold $\tau$:
\begin{equation}
\pi(M_{t,i}, M^{\text{hand}}_{t}) \;=\;
\mathbf{1}\left\{
\frac{|M_{t,i} \cap \mathrm{dilate}(M^{\text{hand}}_{t}, r)|}{|M_{t,i}|}
\ge \tau
\right\}.
\label{eq:near-hand}
\end{equation}
This defines a filtered activated set
\begin{equation}
\mathcal{A}^{\text{nh}}_t \;=\; \left\{\, i \in \mathcal{A}_t \;\middle|\;
\pi\!\left(M_{t,i}, M^{\text{hand}}_t\right)=1 \,\right\},
\label{eq:nh-set}
\end{equation}
and we obtain the corresponding object prior by replacing $\mathcal{A}_t$ in Eq.~\eqref{eq:obj-union} with $\mathcal{A}^{\text{nh}}_t$.

Finally, our dynamic activity prior unions hands and (activated) objects:
\begin{equation}
D_t(\mathbf{x}) \;=\; D^{\text{hand}}_t(\mathbf{x}) \,\vee\, D^{\text{obj}}_t(\mathbf{x}).
\label{eq:interaction-gating}
\end{equation}

We export $\{D_t\}$ as strict binary masks (0/255) for visualization and for
integration into reconstruction.

\begin{figure}[t]
    \centering
    \includegraphics[width=\linewidth]{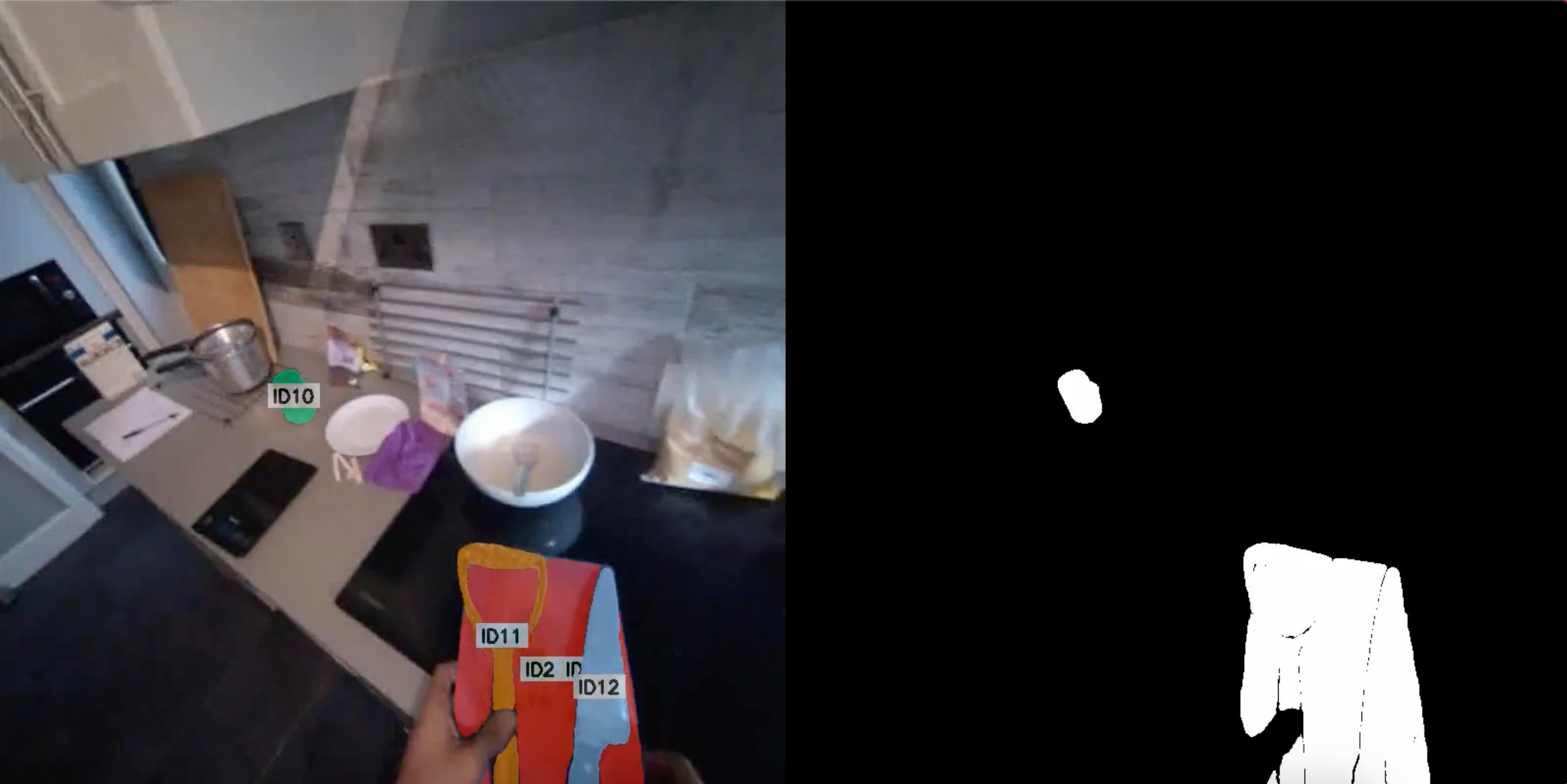}
    \caption{\textbf{Interaction-activated dynamic prior.} We show a SAM3
    tracking frame with instance IDs together with the corresponding
    interaction-activated binary prior $D_t$ (hands always masked; objects
    masked cumulatively after onset), which is used to construct attention masks
    in the reconstruction module.}
    \label{fig:dynamic-prior-vis}
\end{figure}

\subsection{Module B: Chunk-wise Reconstruction with Token-Level Dynamic Gating}
\label{sec:method-recon}

For each temporal chunk $c$, we run MapAnything to obtain per-frame depth, intrinsics (estimated internally), chunk-local camera poses
$\{\hat{T}^{(c)}_t\}$, and a dense point cloud $\hat{\mathcal{P}}^{(c)}$. To prevent dynamic content (hands and manipulated objects) from contaminating the static reconstruction, we inject the interaction-aware prior $D_t$ (\S\ref{sec:method-dynamic-prior}) as an \emph{attention mask over 2D image tokens} during inference. This is an inference-time modification (no retraining or parameter updates).

\paragraph{B1: 2D token grid and pixel-to-token masking.}
MapAnything represents each frame using a 2D grid of image tokens produced by a DINOv2-style patch tokenizer. Let $I_t$ be transformed (resize/crop/pad) into the tokenizer input resolution $H'\times W'$ as part of the standard MapAnything preprocessing; we apply the \emph{same} geometric transform to the binary pixel mask $D_t$ to obtain $\tilde{D}_t\in\{0,1\}^{H'\times W'}$. Let the tokenizer
partition $\tilde{D}_t$ into a token grid with patch size $P$ and cell index $(u,v)$. Define the pixel set covered by each cell:
\begin{equation}
\Omega(u,v)=\{(x,y)\mid uP \le x < (u{+}1)P,\;\; vP \le y < (v{+}1)P\},
\end{equation}
and compute a token-level indicator by max-pooling:
\begin{equation}
m_{t,(u,v)} \;=\; \max_{(x,y)\in \Omega(u,v)} \tilde{D}_t(x,y).
\label{eq:pixel2token}
\end{equation}
Thus a token is masked if any pixel in its covered region is marked dynamic. In our experiments, MapAnything uses a fixed $H'{=}W'{=}800$ and patch size $P{=}14$, but the construction above applies to any tokenizer resolution.

\begin{figure}[t]
  \centering
  \includegraphics[width=\linewidth]{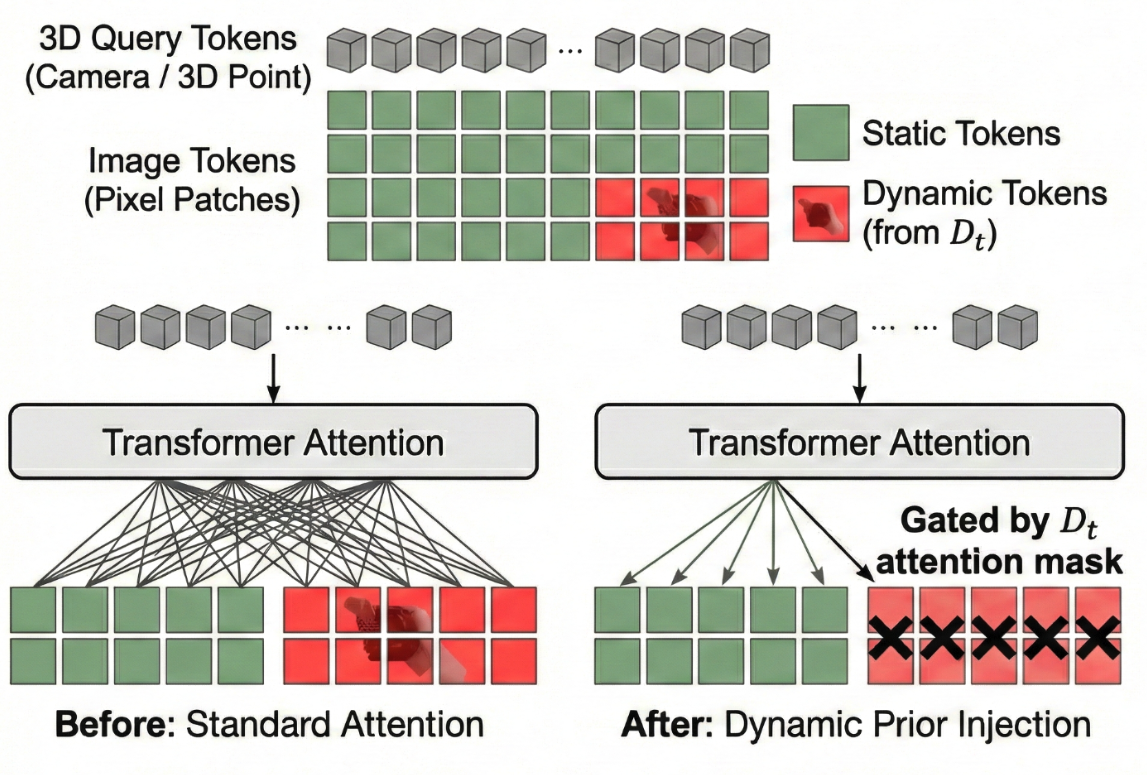}
  \caption{\textbf{Dynamic prior injection for robust 3D reconstruction.} 
  We augment the standard transformer attention between 3D query tokens (cameras / 3D points) and image tokens (pixel patches) by gating cross-attention with a dynamic mask $D_t$. Tokens marked as dynamic (red) are suppressed, so 3D queries attend primarily to static regions (green), reducing motion-induced contamination and ghosting artifacts in the reconstructed scene.}
  \label{fig:attention_mech}
\end{figure}

\paragraph{B2: Attention masking in MapAnything.}
MapAnything updates 2D image tokens via temporal 2D self-attention and updates 3D latent tokens via 2D$\rightarrow$3D cross-attention. Let an attention module have logits $\alpha_{ij}=\frac{q_i^\top k_j}{\sqrt{d}}$ and weights $A_{ij}=\mathrm{softmax}_j(\alpha_{ij})$. We implement dynamic suppression by adding an infinite negative bias to masked \emph{keys/values}:
\begin{equation}
b_j=
\begin{cases}
0,& m_j=0\\
-\infty,& m_j=1
\end{cases},
\quad
\tilde{\alpha}_{ij}=\alpha_{ij}+b_j,
\label{eq:attn_mask}
\end{equation}

\begin{equation}
A_{ij}=\mathrm{softmax}_j(\tilde{\alpha}_{ij}),
\label{eq:logits_output}
\end{equation}
which guarantees $A_{ij}=0$ for masked tokens after the softmax. We apply this masking (i) in temporal 2D self-attention to prevent dynamic evidence from propagating across frames, and (ii) in 2D$\rightarrow$3D cross-attention so that dynamic 2D tokens cannot update 3D tokens as shown in Fig.~\ref{fig:attention_mech}. Consequently, the reconstructed depth, poses, and dense point clouds are driven primarily by static regions.

\begin{figure*}[t]
    \centering
    \includegraphics[width=\textwidth]{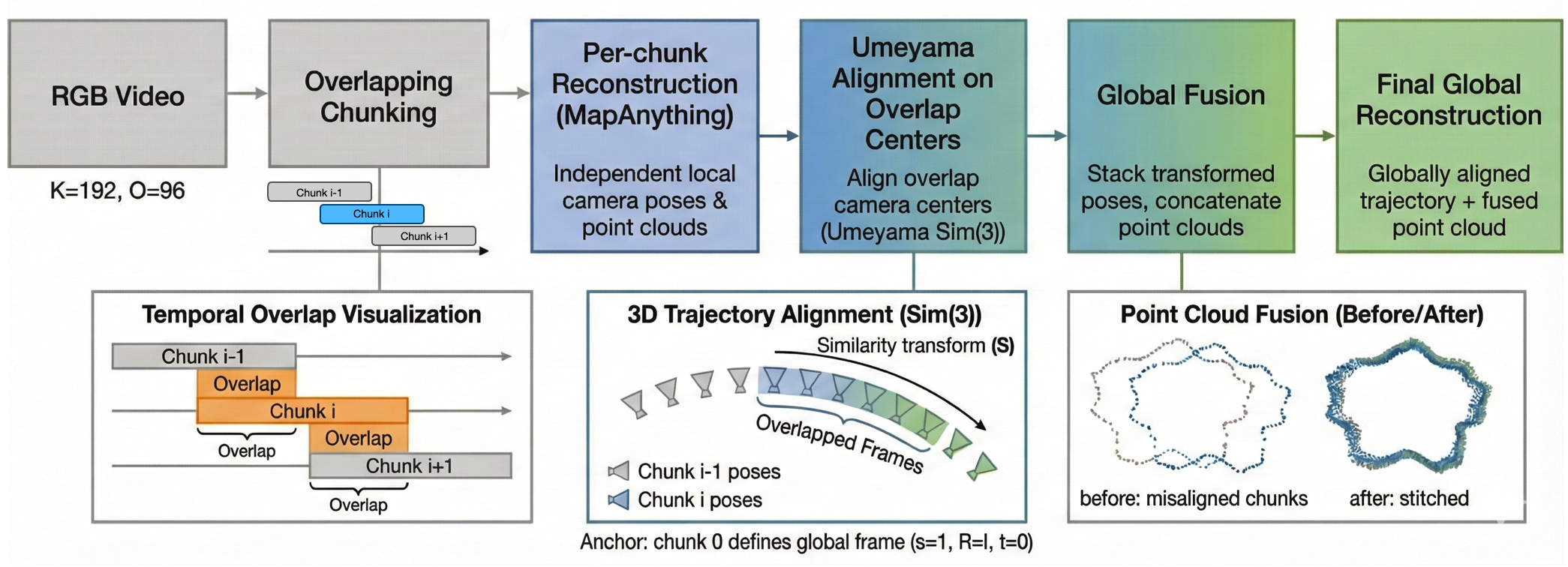}
    \caption{\textbf{Module B3\&4: chunk-wise reconstruction and overlap-based Sim(3) stitching.}
    We split a long egocentric RGB video into overlapping temporal chunks (here $K{=}192$ frames with overlap $O{=}96$).
    Each chunk is reconstructed independently by MapAnything, producing chunk-local camera poses and dense point clouds.
    To obtain a globally aligned trajectory and static map, we estimate a similarity transform between consecutive chunks
    by Umeyama alignment on \emph{overlap camera centers}, anchoring the global frame with the first chunk ($s{=}1, R{=}I, \mathbf{t}{=}\mathbf{0}$).
    We then apply the estimated Sim(3) transform to map each chunk into the global frame and fuse transformed point clouds
    (with optional subsampling for memory control), yielding a stitched global camera trajectory and a fused point cloud with reduced cross-chunk misalignment.}
    \label{fig:moduleB-stitching}
\end{figure*}

\paragraph{B3: Pose convention and overlap camera centers.}
Because MapAnything is not a SLAM system, chunk-local reconstructions may lie in independent coordinate frames and can exhibit scale drift. We therefore stitch chunks sequentially using overlap-based Sim(3) alignment on the \emph{overlap trajectory} as shown in Fig.~\ref{fig:moduleB-stitching}. 

For each chunk $c$, MapAnything outputs per-frame \emph{camera-to-world} poses 
\begin{equation}
\hat{T}^{(c)}_t=[\hat{R}^{(c)}_t \mid \hat{\mathbf{t}}^{(c)}_t]\in SE(3)
\end{equation}
for frames $t$ in the chunk. Under this convention, the camera center in the chunk coordinate frame is simply the translation component:
\begin{equation}
\hat{\mathbf{c}}^{(c)}_t \;=\; \hat{\mathbf{t}}^{(c)}_t \in \mathbb{R}^3.
\label{eq:cam-center}
\end{equation}
Let chunk $c$ overlap with chunk $c{-}1$ on $L$ frames with indices $\mathcal{O}_c=\{t_1,\dots,t_L\}$. Stacking overlap centers yields $\mathbf{X}\in\mathbb{R}^{L\times 3}$ for the current chunk, $\mathbf{X}_k=\hat{\mathbf{c}}^{(c)}_{t_k}$, and $\mathbf{Y}\in\mathbb{R}^{L\times 3}$ for the already-stitched global trajectory at the same frame indices.

\paragraph{B4: Umeyama Sim(3) alignment and scale anchoring.}
We estimate a similarity transform $S_c=(s_c,R_c,\mathbf{t}_c)\in\mathrm{Sim}(3)$ by Umeyama:
\begin{equation}
\min_{s,R,\mathbf{t}} \sum_{k=1}^{L}
\left\|\mathbf{Y}_k - (s R \mathbf{X}_k + \mathbf{t})\right\|_2^2, R\in SO(3),\ s>0.
\label{eq:umeyama}
\end{equation}
We anchor the global frame by setting the first chunk as the origin and scale: $S_1=(1,I,\mathbf{0})$. For $c\ge 2$, we map the chunk-$c$ camera \emph{centers} and 3D points into the global frame using $S_c$:
\begin{equation}
\mathbf{c}_t \;=\; s_c R_c \hat{\mathbf{c}}^{(c)}_t + \mathbf{t}_c,\quad
\mathcal{P}\;:=\;\bigcup_c S_c(\hat{\mathcal{P}}^{(c)}),
\end{equation}
thereby maintaining a consistent global scale across stitched chunks. In practice, we apply the same $(s_c,R_c,\mathbf{t}_c)$ to all chunk-local points, and compose the chunk-local poses with $(R_c,\mathbf{t}_c)$ while applying $s_c$ to the translational component.\footnote{Equivalently, $s_c$ acts on 3D points and camera positions, while pose rotations remain in $SO(3)$.}
We aggregate transformed dense point clouds across chunks with subsampling to control memory, yielding a stitched global camera trajectory and a global static point cloud.

\subsection{Outputs}
\label{sec:method-outputs}

The pipeline produces (i) per-frame binary dynamic priors $\{D_t\}$ (hands always masked; objects masked cumulatively after onset), (ii) chunk-wise reconstructions mapped into a global frame after overlap-based Umeyama Sim(3) stitching, and (iii) an aggregated dense \emph{point cloud} for downstream tasks.

\section{Experiments}
\label{sec:experiments}

We evaluate whether (i) overlap-based Sim(3) stitching produces a stable
\emph{globally aligned} trajectory and point cloud from chunk-local reconstructions, and
(ii) interaction-aware dynamic priors injected as \emph{token-level attention masks}
reduce dynamic contamination (ghosting) in the static map. During development we found
that several seemingly reasonable “ghosting” proxies can be dominated by depth
\emph{completeness} effects. We therefore report (a) trajectory/overlap consistency,
(b) depth coverage, and (c) \emph{density-normalized} dynamic contamination metrics, together
with qualitative point-cloud comparisons.

\subsection{Experimental Setup}
\label{sec:exp-setup}

\paragraph{Data.}
We evaluate on long egocentric sequences from EPIC-HD/HD-EPIC.
We report results on sequence \textbf{20240414-172534} (participant \textbf{P04}),
containing \textbf{3565} frames at \textbf{30} FPS. Hands and manipulated objects appear
frequently throughout the sequence.

\paragraph{Reconstruction backbone and chunking.}
We use MapAnything as the reconstruction backbone.
Due to GPU memory/runtime limits of dense-token temporal attention, we run inference in
overlapping temporal chunks.
For the 30\,FPS run (\texttt{full\_len}), we use chunk length $K{=}180$ with overlap
$O{=}90$ frames, producing \textbf{40} chunks over \textbf{3565} frames.
For most ablations we downsample temporally to \textbf{6\,Hz} (keep every 5th frame),
yielding \textbf{713} frames; with the same $K{=}180,O{=}90$ we obtain \textbf{8} chunks.
Each chunk outputs per-frame depth, intrinsics (estimated internally),
chunk-local poses $\{\hat{T}^{(c)}_t\}$, and a dense point cloud $\hat{\mathcal{P}}^{(c)}$.
We stitch chunks sequentially via overlap-based Umeyama alignment on camera centers.

\paragraph{Interaction-aware dynamic priors and mask variants.}
We follow \S\ref{sec:method-dynamic-prior}. Briefly, a SymSkill-style
interaction/skill segmentor partitions the video into short clips and outputs SPO predictions.
An LLM selects motion-inducing predicates to identify candidate manipulated objects and their
interaction onsets. A VLM provides coarse localization cues (tags/boxes) used to initialize SAM3,
which tracks instances and produces per-frame masks.

We consider two \emph{suppression mask} strategies during reconstruction:
(i) \textbf{dynamic-only} masking, which suppresses only pixels predicted dynamic at time $t$,
and (ii) \textbf{cumulative / full-time} masking, which keeps an object masked after its first
interaction onset (our intended design in \S\ref{sec:method-dynamic-prior}).

\paragraph{Evaluation masks: instantaneous vs. footprint.}
For evaluation (metric conditioning), we use two binary mask sets:
\texttt{union\_mask\_dynamics} (per-frame instantaneous dynamic regions) and
\texttt{union\_mask\_fulltime} (dynamic footprint / full-time union).
The latter is useful to avoid counting “stale” ghost geometry (left behind by moved objects)
as belonging to the static region.

Unless stated otherwise, we condition \emph{static-region} metrics (e.g., $B_{\text{static}}$,
$D_{\text{static}}$) on \texttt{union\_mask\_fulltime} to ensure that any pixel ever occupied
by hands/objects is excluded from the static evaluation region.

\paragraph{Implementation details.}
All experiments run on \textbf{RTX-5090} with \textbf{32}\,GB.
Average runtime is \textbf{49}\,sec/chunk and \textbf{6}\,min \textbf{13}\,sec/sequence
(including stitching) in the 6\,Hz setting.


\subsection{Metrics}
\label{sec:exp-metrics}

We report both stitching/trajectory consistency metrics and reconstruction-level contamination metrics.
All metrics are computed from per-chunk raw outputs (depth, intrinsics, poses), without relying on the
final globally fused point cloud.

\paragraph{Pose convention and camera centers.}
MapAnything outputs per-frame poses as camera-to-world transforms
$T=\begin{bmatrix}R & \mathbf{t}\\ \mathbf{0}^\top & 1\end{bmatrix}$, i.e.,
$\mathbf{p}_w = R\,\mathbf{p}_c + \mathbf{t}$.
We define the camera center as $\mathbf{c}=\mathbf{t}$ and use it consistently for overlap alignment
and trajectory metrics.

\paragraph{(A) Multi-surface ratio (auxiliary ghosting proxy).}
We keep the multi-surface ratio $\rho$ (Eq.~(19) in \S\ref{sec:exp-metrics} of the original draft) as an auxiliary proxy,
computed with (i) visibility gating ($z_{\text{proj}}\le z_{\text{depth}}+\varepsilon$, with $\varepsilon{=}0.05$),
(ii) mask dilation radius 3 for region conditioning, and (iii) a $\texttt{cnt}\ge 2$ denominator to reduce single-hit noise.
Empirically, $\rho$ can be dominated by depth holes and is not always aligned with qualitative ghosting; we therefore do not use it
as the primary quantitative evidence of contamination reduction.

\paragraph{(B) Overlap geometry consistency (primary overlap metric).}
For overlap frame $t\in\mathcal{O}_{c+1}$, let $\mathcal{Q}^{(c)}_t$ and $\mathcal{Q}^{(c+1)}_t$
be back-projected point sets from depth in chunks $c$ and $c{+}1$.
We map the latter into chunk-$c$ coordinates by $\widetilde{\mathcal{Q}}^{(c+1)}_t := S_{c+1}(\mathcal{Q}^{(c+1)}_t)$.
Define
\begin{equation}
d(\mathbf{x},\mathcal{A}) \;:=\; \min_{\mathbf{a}\in\mathcal{A}} \|\mathbf{x}-\mathbf{a}\|_2 .
\label{eq:nn-dist}
\end{equation}
and compute a symmetric nearest-neighbor distance:
\begin{equation}
d^{\text{geo}}_t \;=\; \frac{1}{2}\left(
\frac{1}{|\widetilde{\mathcal{Q}}^{(c+1)}_t|}
\sum_{\mathbf{q}\in\widetilde{\mathcal{Q}}^{(c+1)}_t} d(\mathbf{q},\mathcal{Q}^{(c)}_t)
+
\frac{1}{|\mathcal{Q}^{(c)}_t|}
\sum_{\mathbf{p}\in\mathcal{Q}^{(c)}_t} d(\mathbf{p},\widetilde{\mathcal{Q}}^{(c+1)}_t)
\right).
\label{eq:overlap-geo}
\end{equation}
We report $B_{\text{all}}=\mathbb{E}[d^{\text{geo}}_t]$.
We also report $B_{\text{static}}$ by building $\mathcal{Q}_t$ only from pixels where the evaluation mask is static.

\paragraph{(C) Density-normalized dynamic contamination (primary ghosting metric).}
To avoid being dominated by total point coverage, we introduce \emph{density-style} contamination metrics.
Given a chunk-level point cloud $\mathcal{P}^{(c)}$ (subsampled) projected into a frame $t$, define:
\begin{itemize}
\item $N_{\text{dyn}}(t)$: number of projected points landing in dynamic region $\Omega_{\text{dyn}}(t)$,
\item $H_{\text{dyn}}(t)$: number of dynamic pixels hit by at least one projected point,
\item similarly $N_{\text{static}}(t)$ and $H_{\text{static}}(t)$ on $\Omega_{\text{static}}(t)$.
\end{itemize}
We report three normalized ratios (dyn/static):
\begin{align}
C^{\text{den}}_{\text{ratio}} &= \frac{\mathbb{E}\left[\frac{N_{\text{dyn}}(t)}{|\Omega_{\text{dyn}}(t)|}\right]}
{\mathbb{E}\left[\frac{N_{\text{static}}(t)}{|\Omega_{\text{static}}(t)|}\right]}, \\
C^{\text{occ}}_{\text{ratio}} &= \frac{\mathbb{E}\left[\frac{H_{\text{dyn}}(t)}{|\Omega_{\text{dyn}}(t)|}\right]}
{\mathbb{E}\left[\frac{H_{\text{static}}(t)}{|\Omega_{\text{static}}(t)|}\right]}, \\
C^{\text{od}}_{\text{ratio}} &= \frac{\mathbb{E}\left[\frac{N_{\text{dyn}}(t)}{H_{\text{dyn}}(t)+\epsilon}\right]}
{\mathbb{E}\left[\frac{N_{\text{static}}(t)}{H_{\text{static}}(t)+\epsilon}\right]},
\end{align}
where $C^{\text{den}}_{\text{ratio}}$ is the main “points-per-pixel” contamination density,
$C^{\text{occ}}_{\text{ratio}}$ is a “hit-rate” occupancy proxy, and $C^{\text{od}}_{\text{ratio}}$ measures overdraw / thickness.
All are \emph{lower-is-better} (less dynamic contamination relative to static).

\paragraph{Scale stability.}
Umeyama returns a per-transition scale $s_c$. We report the mean scale $\bar{s}$ as a drift proxy
(near 1 indicates stable scale across chunk transitions).

\paragraph{(D) Depth coverage / completeness.}
A key confound we found is that more aggressive masking can reduce depth completeness, which in turn can
inflate overlap geometry errors and can also change ray-based “ghosting” proxies.
We therefore report valid-depth ratios computed directly from predicted depth and the evaluation masks:
\begin{align}
D_{\text{all}} &= \frac{\#\{(u,v): z(u,v)\ \text{finite and } z(u,v)>0\}}{HW}, \\
D_{\text{dyn}} &= \frac{\#\{(u,v)\in\Omega_{\text{dyn}}: z(u,v)\ \text{valid}\}}{|\Omega_{\text{dyn}}|}, \\
D_{\text{static}} &= \frac{\#\{(u,v)\in\Omega_{\text{static}}: z(u,v)\ \text{valid}\}}{|\Omega_{\text{static}}|}.
\end{align}

\paragraph{(E) Overlap camera-center residual (trajectory sanity check).}
For each chunk transition $c\rightarrow c{+}1$, let $\mathcal{O}_{c+1}$ denote overlap frames.
After estimating $S_{c+1}\in \mathrm{Sim}(3)$ by Umeyama on overlap camera centers, we compute:
\begin{equation}
e^{\text{cen}}_{c+1}
\;=\;
\sqrt{\frac{1}{|\mathcal{O}_{c+1}|}\sum_{t\in\mathcal{O}_{c+1}}
\left\|\mathbf{c}^{(c)}_t - S_{c+1}\big(\mathbf{c}^{(c+1)}_t\big)\right\|_2^2}.
\label{eq:center-rms}
\end{equation}
We report the mean/median over all transitions.

\subsection{Baselines and Ablations}
\label{sec:exp-baselines}

We evaluate four variants:

\begin{itemize}
    \item \textbf{MapAnything (baseline)}:
    chunk-wise reconstruction with Sim(3) stitching, \emph{no} dynamic suppression.

    \item \textbf{Naive blackout (pixel-level)}:
    blank masked pixels in the input frames using the same mask set as the suppression prior, without modifying attention.

    \item \textbf{Attention gating (dynamic-only)}:
    inject per-frame dynamic masks as a token-level attention mask (Eq.~(10-11) in \S\ref{sec:method-recon}).

    \item \textbf{Attention gating (cumulative / full-time) (ours)}:
    inject interaction-activated cumulative masks (objects remain masked after onset), also as token-level attention masks.
\end{itemize}

\begin{table*}[t]
\centering
\small
\begin{tabular}{@{}lccc@{}}
\toprule
\textbf{Variant} & \textbf{Suppression} & \textbf{Mask strategy} & \textbf{Stitching} \\
\midrule
Baseline & none & - & Sim(3) \\
Blackout & pixel & dynamic-only / cumulative & Sim(3) \\
Attn gating (dynamic-only) & attention & per-frame & Sim(3) \\
Attn gating (cumulative) & attention & after-onset, persistent & Sim(3) \\
\bottomrule
\end{tabular}
\caption{Summary of evaluated variants.}
\label{tab:variants}
\end{table*}

\subsection{Overlap-Based Stitching and Cross-Chunk Stability}
\label{sec:exp-stitching}

We first report overlap-based stitching metrics in the 6\,Hz setting (8 chunks, 7 overlaps).
Table~\ref{tab:stitching-main} summarizes overlap consistency and completeness.

\begin{table*}[t]
\centering
\small
\begin{tabular}{@{}lcccccc@{}}
\toprule
\textbf{Variant}
& $e^{\text{cen}}{\downarrow}$
& $B_{\text{all}}{\downarrow}$
& $B_{\text{static}}{\downarrow}$
& $D_{\text{all}}{\uparrow}$
& $D_{\text{static}}{\uparrow}$
& $\bar{s}$ \\
\midrule
Baseline
& 0.019497
& 0.035132
& 0.034525
& 0.945419
& 0.952275
& 1.004606 \\
Blackout
& 0.024591
& 0.043171
& 0.042285
& 0.951403
& 0.960077
& 0.998633 \\
Attn gating (dynamic-only)
& 0.051678
& 0.083331
& 0.083431
& 0.800966
& 0.828438
& 0.997084 \\
Attn gating (cumulative) (ours)
& 0.072496
& 0.100158
& 0.096840
& 0.780635
& 0.809198
& 1.003003 \\
\bottomrule
\end{tabular}
\caption{\textbf{Overlap consistency and completeness (6\,Hz, 8 chunks).}
We condition static-region metrics ($B_{\text{static}}$, $D_{\text{static}}$) on the footprint evaluation mask
\texttt{union\_mask\_fulltime} to avoid counting any ever-dynamic pixels as static.
Dynamic suppression increases overlap residuals and geometry discrepancy, and is accompanied by a large drop in depth completeness
($D_{\text{all}}$ from 0.945419 to 0.800966 / 0.780635). Scale remains stable (mean scale $\bar{s}\approx 1$).}
\label{tab:stitching-main}
\end{table*}

\paragraph{Key observation: overlap errors correlate with completeness.}
Compared to baseline, attention gating increases $B_{\text{all}}$ from 0.035132 to 0.083331 (dynamic-only)
and 0.100158 (cumulative), while $D_{\text{all}}$ drops from 0.945419 to 0.800966 and 0.780635.
This indicates that overlap geometry inconsistency is strongly influenced by missing depth/coverage.
The camera-center residual $e^{\text{cen}}$ also increases (0.019497 $\rightarrow$ 0.051678 / 0.072496),
but scale remains near 1, suggesting instability is not dominated by scale drift.

\subsection{Dynamic Priors Reduce Dynamic Contamination (Ghosting)}
\label{sec:exp-dynamic-prior}

\begin{figure*}[t]
    \centering
    \includegraphics[width=\textwidth]{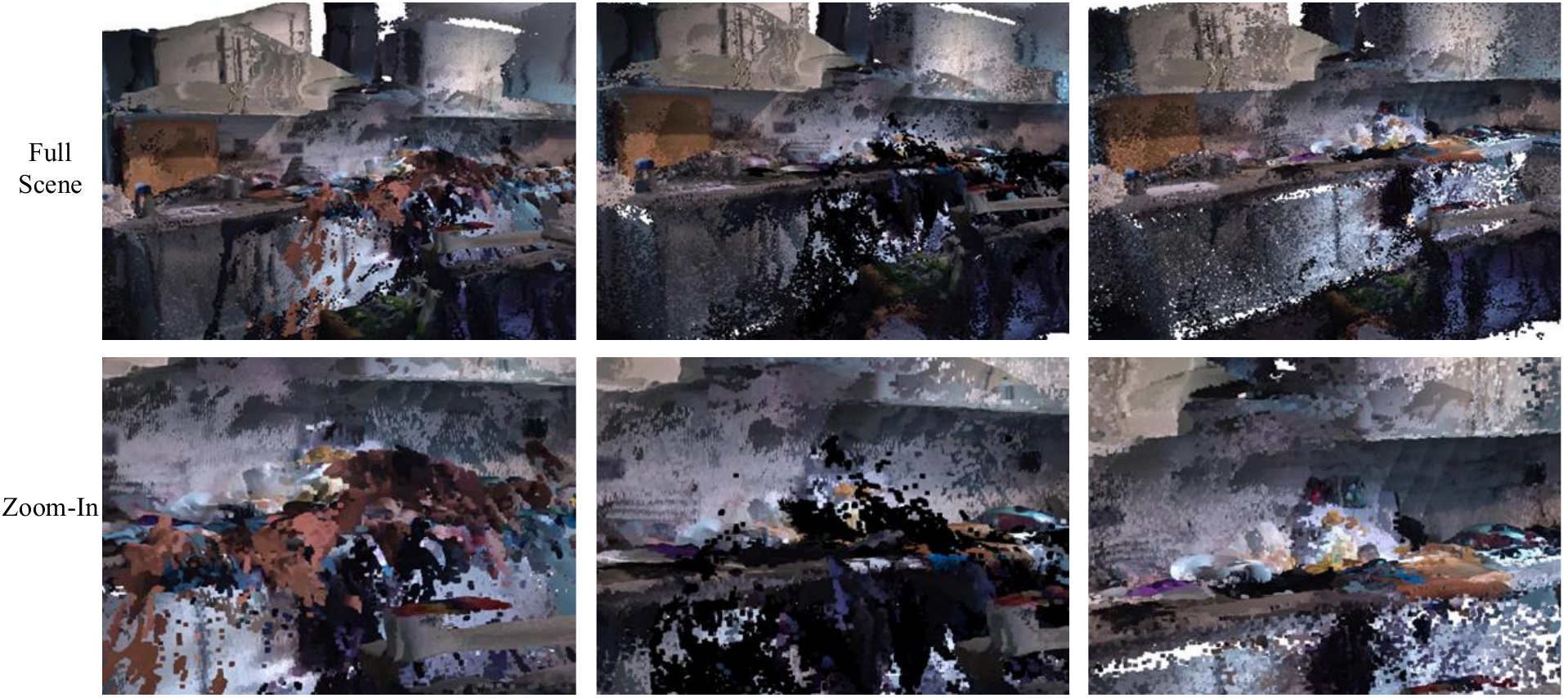}
    \caption{\textbf{Static reconstruction with/without dynamic suppression.}
    Columns (left$\rightarrow$right): baseline MapAnything (no suppression), naive pixel-level blackout, and our token-level attention gating.
    Rows (top$\rightarrow$bottom): full-scene view and a zoom-in on an interaction-heavy region.
    Our method removes dynamic ghosting while better preserving clean static geometry than pixel blackout.
    }
    \label{fig:recon-comparison}
\end{figure*}

\paragraph{Qualitative reconstruction comparison.}
Fig.~\ref{fig:recon-comparison} shows a representative static-map comparison.
Baseline MapAnything exhibits severe ghosting near hands and manipulated objects.
Naive blackout can remove some dynamic texture, but often harms reconstruction completeness and overlap stability.
Token-level attention gating produces visibly cleaner static geometry in the final point cloud, especially in interaction-heavy regions.

\paragraph{Why the original contamination ratio can be misleading.}
A simple “fraction of projected points falling inside the dynamic mask” decreases under masking, but this number
is entangled with overall point/depth coverage.
For example, under footprint evaluation (\texttt{union\_mask\_fulltime}), this ratio decreases from 0.167551 (baseline)
to 0.137780 (dynamic-only gating) and 0.133309 (cumulative gating), but Table~\ref{tab:stitching-main} shows that
depth completeness $D_{\text{all}}$ also drops substantially, which can reduce the numerator by creating holes.

\paragraph{Density-normalized contamination reveals true suppression.}
Table~\ref{tab:contamination-density} reports our density-style contamination metrics under two evaluation masks:
instantaneous dynamic regions (\texttt{union\_mask\_dynamics}) and dynamic footprints (\texttt{union\_mask\_fulltime}).

\begin{table*}[t]
\centering
\small
\begin{tabular}{@{}lccc|ccc@{}}
\toprule
& \multicolumn{3}{c|}{\textbf{Instantaneous mask} (cov=0.077415)}
& \multicolumn{3}{c}{\textbf{Footprint mask} (cov=0.168951)} \\
\textbf{Variant}
& $C^{\text{den}}_{\text{ratio}}{\downarrow}$
& $C^{\text{occ}}_{\text{ratio}}{\downarrow}$
& $C^{\text{od}}_{\text{ratio}}{\downarrow}$
& $C^{\text{den}}_{\text{ratio}}{\downarrow}$
& $C^{\text{occ}}_{\text{ratio}}{\downarrow}$
& $C^{\text{od}}_{\text{ratio}}{\downarrow}$ \\
\midrule
Baseline
& 0.998740 & 1.003221 & 0.988921
& 1.087414 & 1.088884 & 0.997985 \\
Blackout
& 0.992527 & 1.010007 & 0.974469
& 1.072562 & 1.085382 & 0.986884 \\
Attn gating (dynamic-only)
& 0.823669 & 0.848722 & 0.961242
& 0.976011 & 0.988040 & 0.990571 \\
Attn gating (cumulative) (ours)
& 0.807543 & 0.835834 & 0.955473
& 0.963738 & 0.978473 & 0.988066 \\
\bottomrule
\end{tabular}
\caption{\textbf{Density-style contamination (6\,Hz, 8 chunks).}
Lower indicates fewer projected points supported in dynamic regions relative to static regions.
Instantaneous evaluation shows strong suppression: $C^{\text{den}}_{\text{ratio}}$ drops from 0.998740 (baseline)
to 0.823669 (dynamic-only) and 0.807543 (cumulative).
Footprint evaluation (harder, less diluted by “currently static” pixels) still shows consistent reductions
from 1.087414 to 0.976011 and 0.963738.
Naive blackout yields only minor changes compared to attention gating.}
\label{tab:contamination-density}
\end{table*}

\paragraph{Trade-off: cleaner static map vs. reduced completeness and overlap stability.}
Density-normalized contamination improves substantially with attention gating (Table~\ref{tab:contamination-density}),
but completeness decreases (Table~\ref{tab:stitching-main}), which also inflates overlap geometry discrepancy.
This quantitatively explains why $B$ increases despite improved visual static-map purity: overlap Chamfer distances are
sensitive to missing geometry, and camera/geometry overlap constraints weaken when large foreground regions are suppressed.

\paragraph{Multi-surface ratio is not aligned with visual ghosting in this setting.}
With the improved (visibility-gated, static-conditioned) multi-surface metric, the mean multi-surface ratio on the 6\,Hz run is:
$A_{\text{all}}{=}0.067817$ (baseline), $0.046814$ (blackout), but increases to $0.102210$ (dynamic-only gating)
and $0.125036$ (cumulative gating). This contradicts the qualitative reduction in dynamic ghosting, indicating that $\rho$
is dominated by depth incompleteness/holes under masking and should be treated as auxiliary rather than primary evidence.

\subsection{Temporal Downsampling Ablation}
\label{sec:exp-downsample}

\paragraph{Protocol.}
We downsample the 30\,FPS sequence by keeping every 5th frame (30\,Hz $\rightarrow$ 6\,Hz), yielding 713 frames.
We run the same chunked reconstruction and stitching pipeline with $K{=}180,O{=}90$ on the downsampled sequence.
This reduces compute while preserving enough temporal support via overlap.

\paragraph{Notes on comparability.}
Overlap metrics at 6\,Hz vs. 30\,Hz are not directly comparable in absolute value because the temporal baseline between
frames changes; we therefore use downsampling primarily as a practical setting for controlled ablations of masking strategies.

\subsection{Overlap-Based Stitching Improves Global Consistency}
\label{sec:exp-stitching}

We first validate that overlap-based Umeyama stitching is necessary to construct a stable global frame
from chunk-local reconstructions.

\begin{figure}[t]
    \centering
    \begin{subfigure}[b]{0.49\linewidth}
        \centering
        \includegraphics[width=\linewidth]{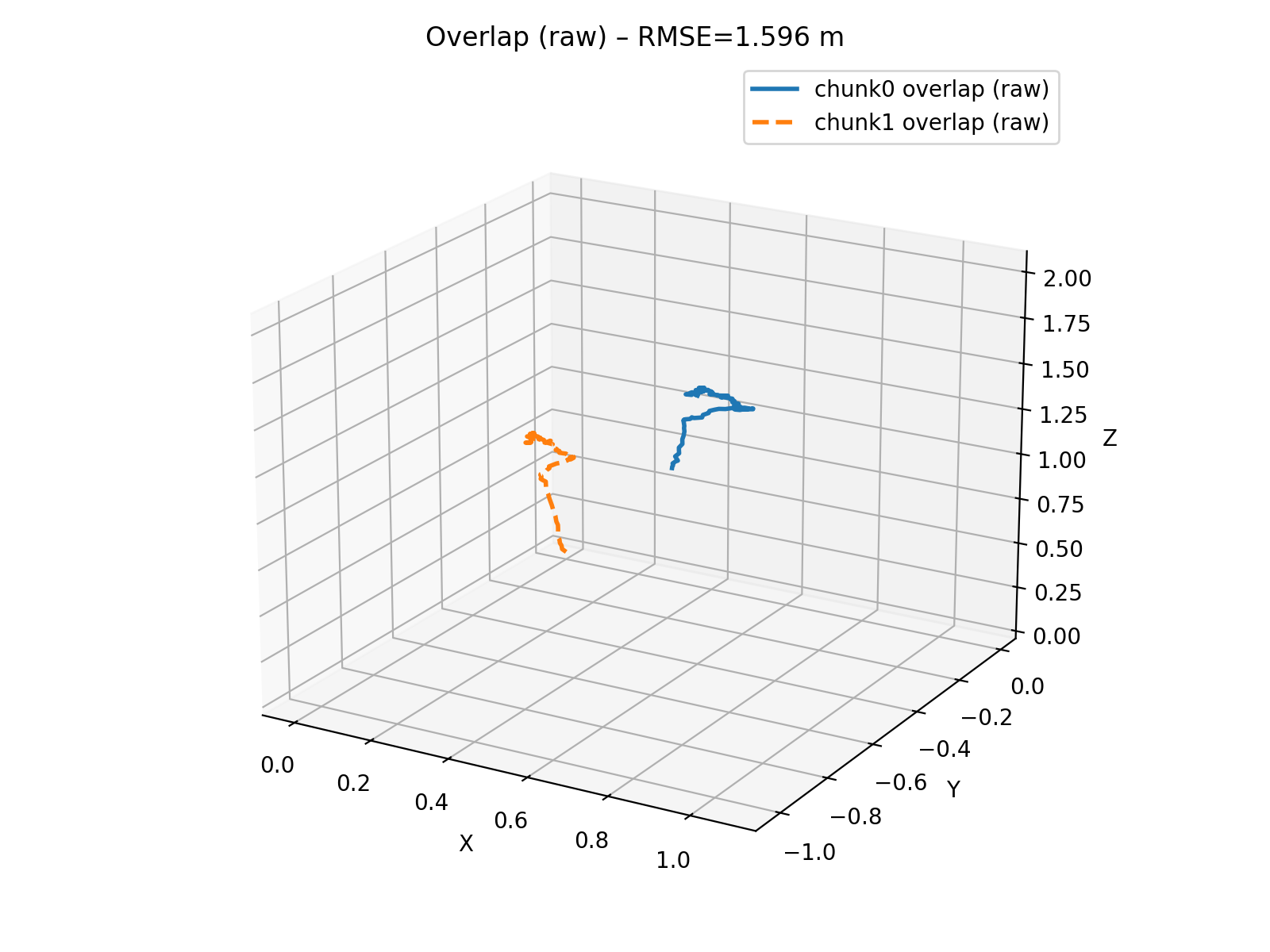}
        \caption{Raw (unstitched) overlap. RMSE \(=1.596\,\mathrm{m}\).}
        \label{fig:overlap-raw}
    \end{subfigure}
    \hfill
    \begin{subfigure}[b]{0.49\linewidth}
        \centering
        \includegraphics[width=\linewidth]{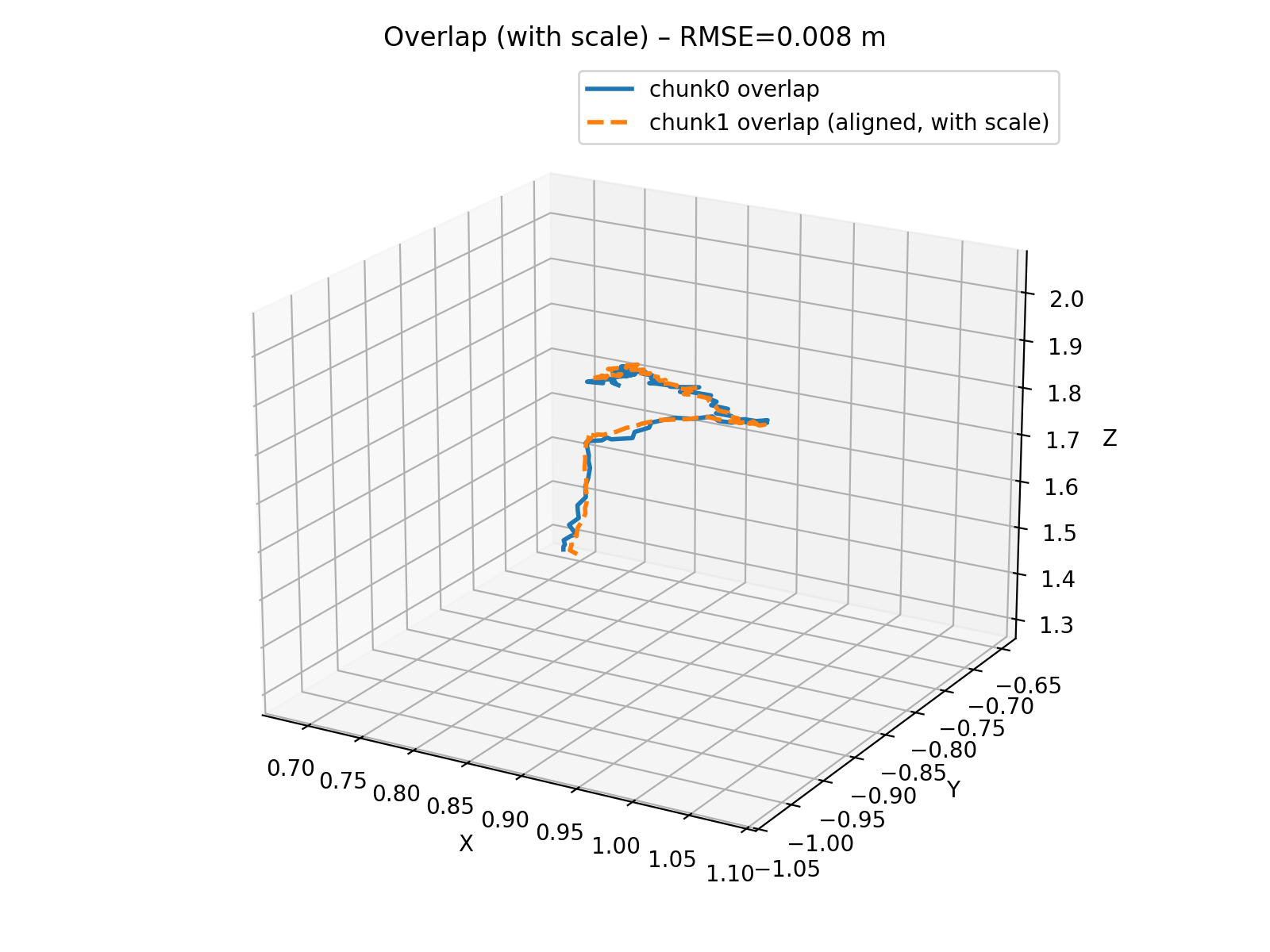}
        \caption{After Umeyama Sim(3) alignment. RMSE \(=0.008\,\mathrm{m}\).}
        \label{fig:overlap-sim3}
    \end{subfigure}
    \caption{\textbf{Overlap alignment for stitching.}
    Camera-center trajectories over the overlap between two adjacent chunks.
    Left: raw (unstitched). Right: after Umeyama Sim(3) alignment with scale.
    }
    \label{fig:overlap-umeyama}
\end{figure}

\paragraph{Overlap residual before/after Sim(3) alignment.}
We visualize camera-center trajectories over the temporal overlap for two adjacent chunks,
before and after Umeyama Sim(3) alignment.
As shown in \cref{fig:overlap-umeyama}, raw chunk-local trajectories can disagree substantially in the overlap
(\( \text{RMSE}=1.596\,\mathrm{m} \)), while Sim(3) alignment (allowing scale) yields tight agreement
(\( \text{RMSE}=0.008\,\mathrm{m} \)).
This supports our choice of Sim(3) stitching when monocular chunks exhibit mild scale inconsistency.


\subsection{Failed Experiments and Lessons Learned}
\label{sec:exp-failures}

\paragraph{Metric pitfall: coverage-dominated “ghosting” proxies.}
A major failure mode in our initial evaluation was that several ghosting proxies were confounded by coverage.
In particular, overlap geometry consistency $B$ can “explode” when depth completeness drops, even if the remaining
geometry is cleaner; similarly, multi-surface ratios can change due to holes and reduced point support rather than
true ghosting reduction. This motivated adding (D) depth coverage and (C) density-normalized contamination.

\paragraph{Tracking without interaction activation.}
We initially attempted long-horizon semantic instance tracking and post-hoc ID merging.
Occlusions and viewpoint changes caused ID fragmentation and drift, making instance-level supervision brittle for long videos.

\paragraph{Lesson.}
For static reconstruction, perfect long-horizon instance tracking is unnecessary.
Instead, reliable identification of regions that become dynamic \emph{after interaction onset} is sufficient,
and injecting this prior directly in attention provides a controllable way to suppress dynamic evidence.

\begin{figure*}[t]
  \centering

  \begin{subfigure}[t]{0.24\textwidth}\centering
    \textbf{Baseline}
  \end{subfigure}\hfill
  \begin{subfigure}[t]{0.24\textwidth}\centering
    \textbf{Blackout}
  \end{subfigure}\hfill
  \begin{subfigure}[t]{0.24\textwidth}\centering
    \textbf{Dynamin (ours)}
  \end{subfigure}\hfill
  \begin{subfigure}[t]{0.24\textwidth}\centering
    \textbf{Overall-mask}
  \end{subfigure}

  \vspace{0.6ex}

  \begin{subfigure}[t]{0.24\textwidth}\centering
    \includegraphics[width=\linewidth]{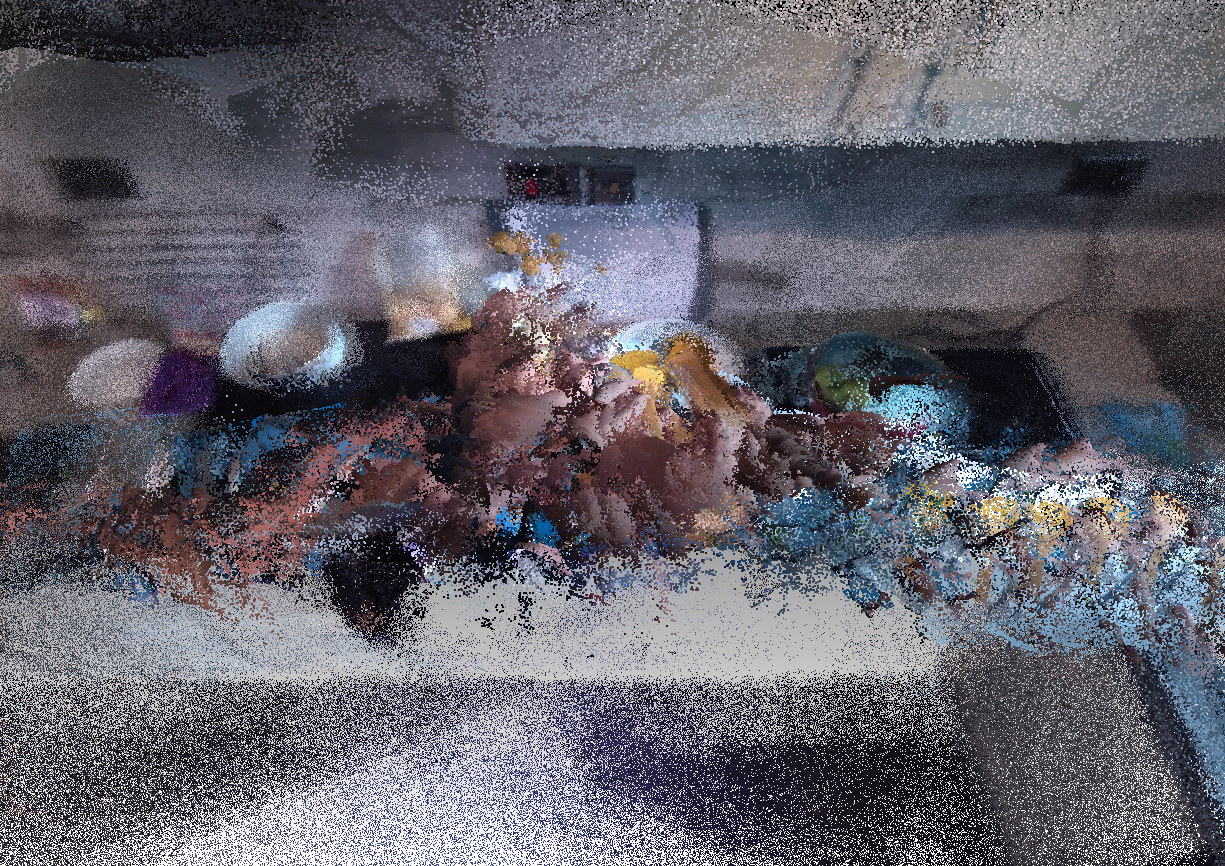}
  \end{subfigure}\hfill
  \begin{subfigure}[t]{0.24\textwidth}\centering
    \includegraphics[width=\linewidth]{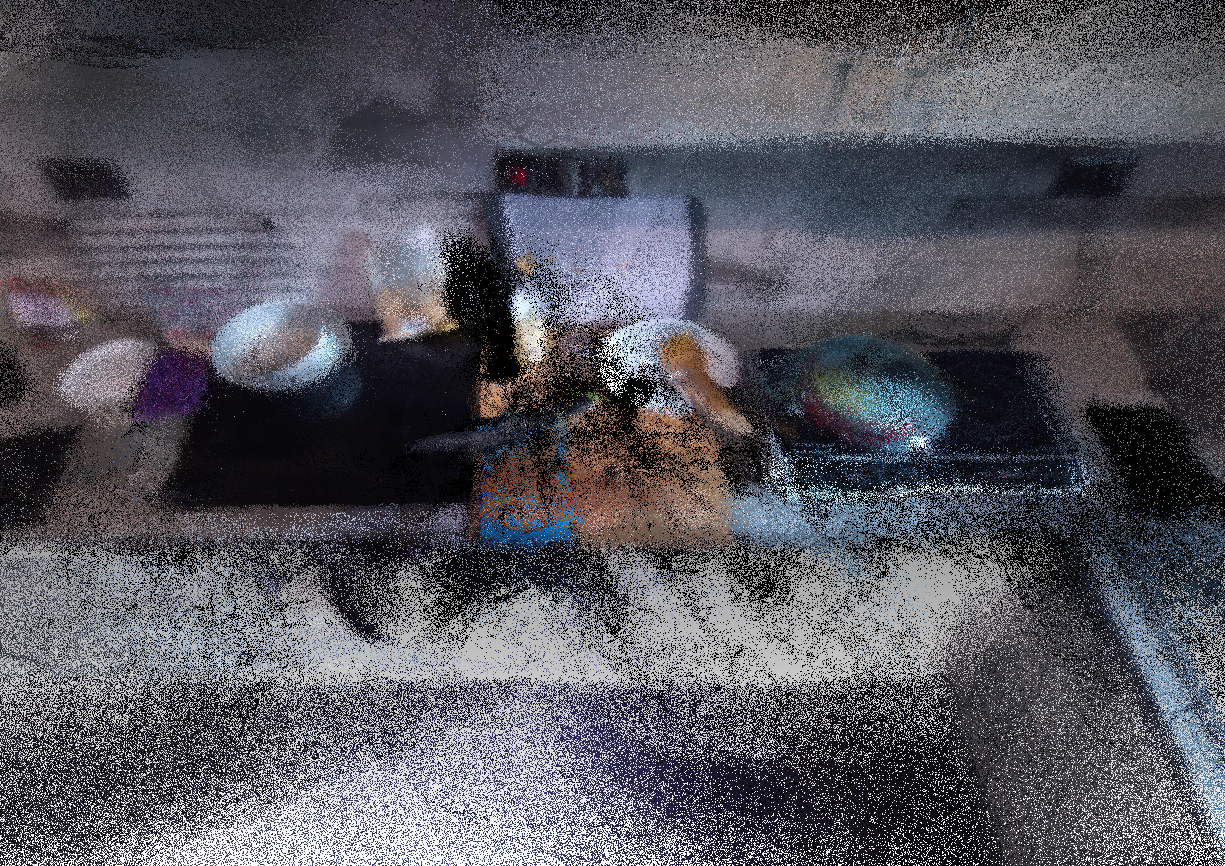}
  \end{subfigure}\hfill
  \begin{subfigure}[t]{0.24\textwidth}\centering
    \includegraphics[width=\linewidth]{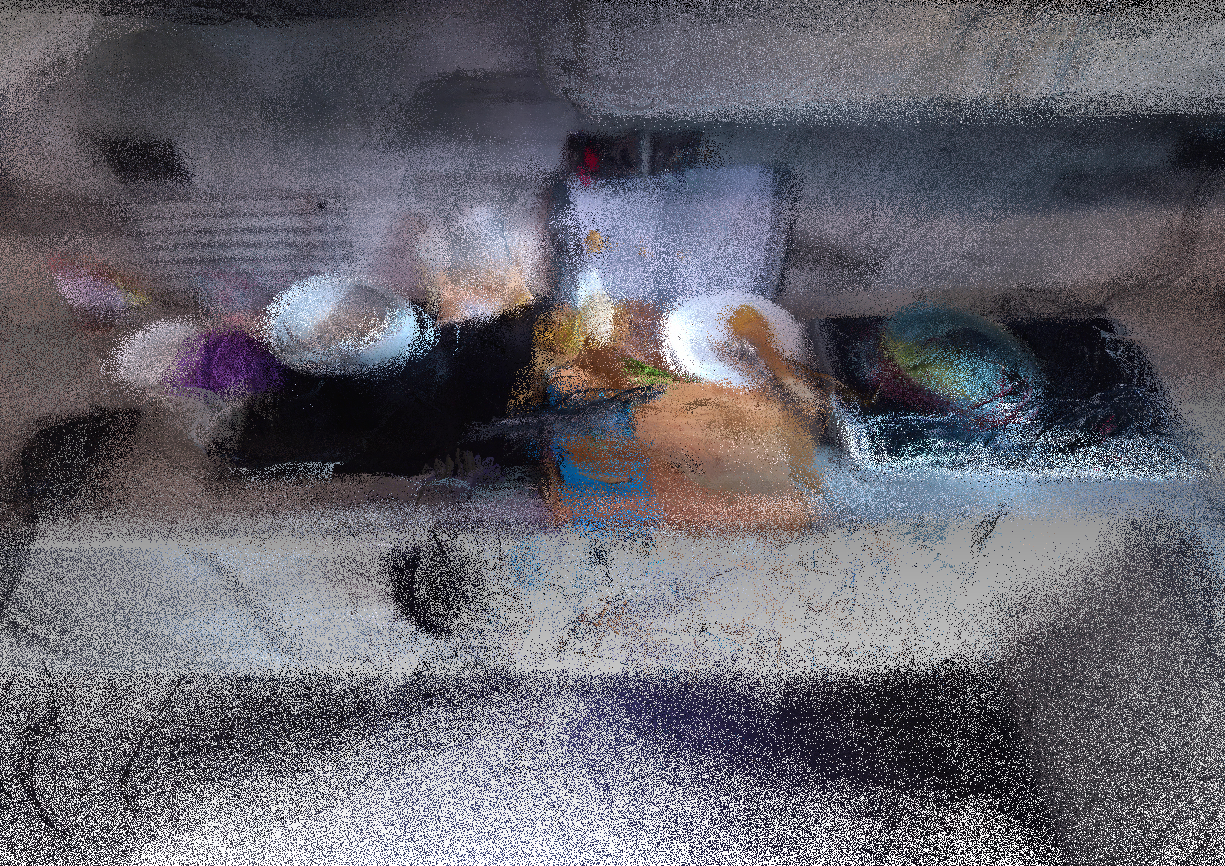}
  \end{subfigure}\hfill
  \begin{subfigure}[t]{0.24\textwidth}\centering
    \includegraphics[width=\linewidth]{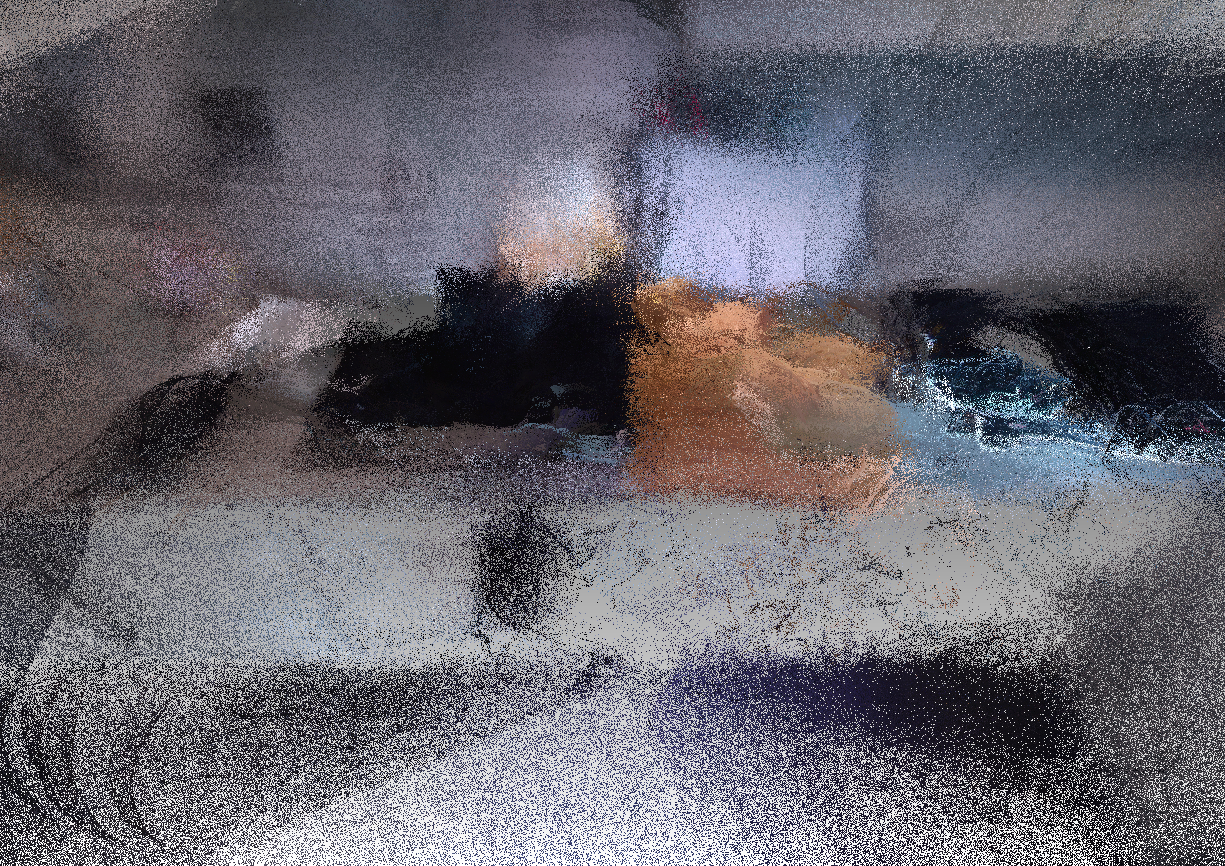}
  \end{subfigure}

  \vspace{0.8ex}

  \begin{subfigure}[t]{0.24\textwidth}\centering
    \includegraphics[width=\linewidth]{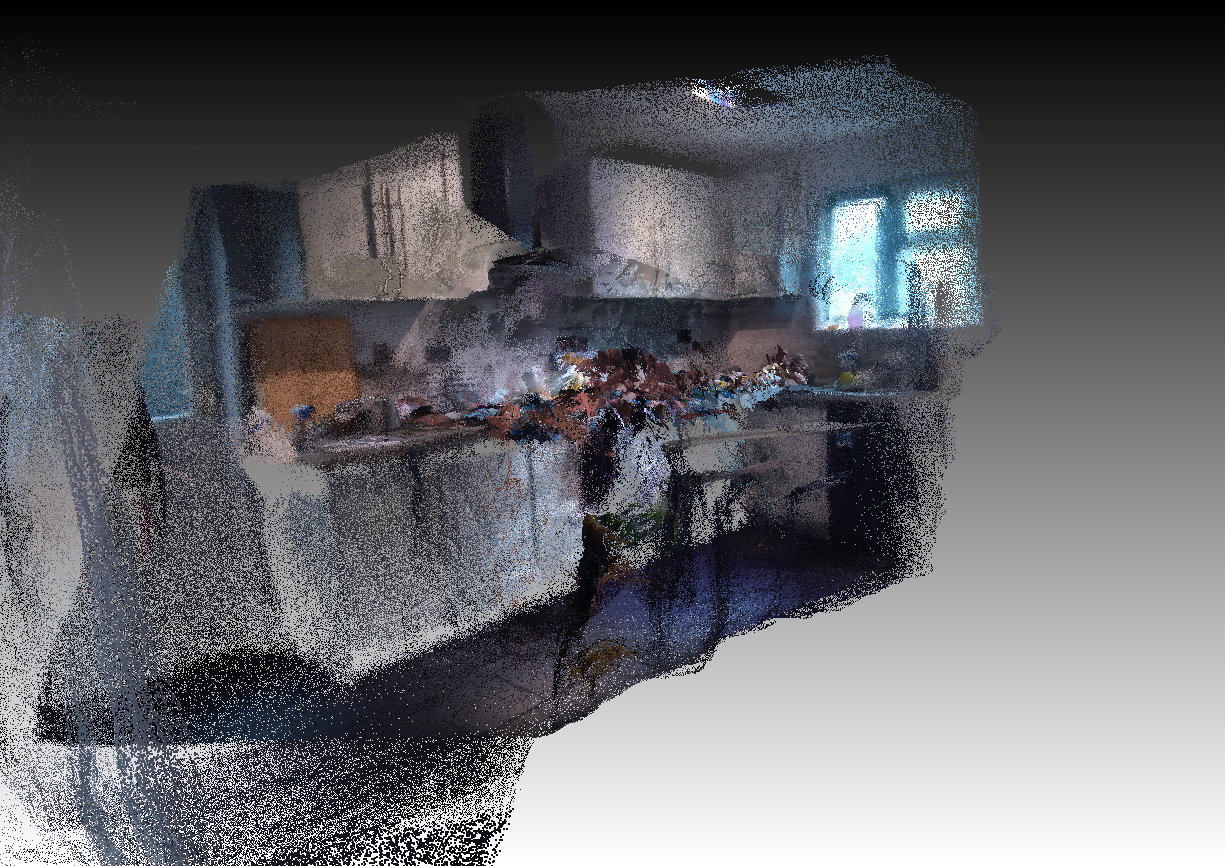}
  \end{subfigure}\hfill
  \begin{subfigure}[t]{0.24\textwidth}\centering
    \includegraphics[width=\linewidth]{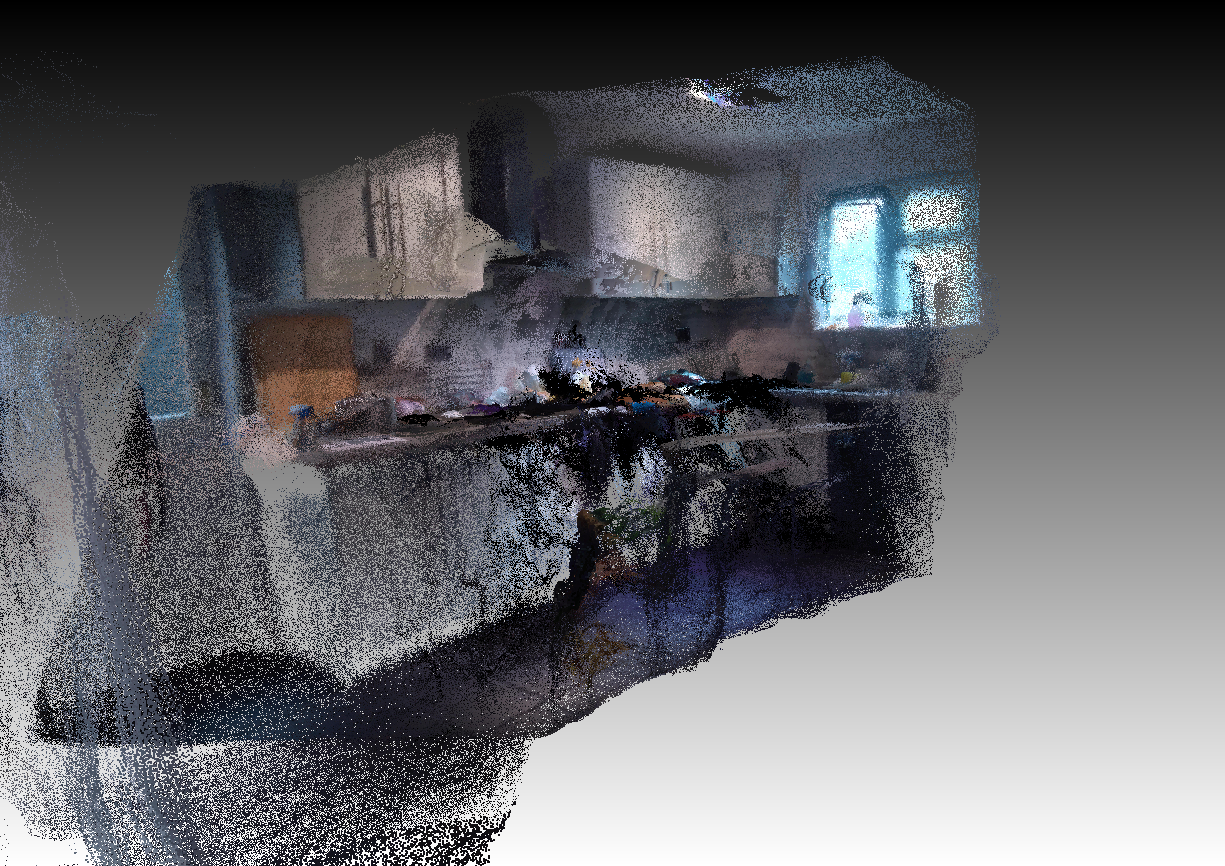}
  \end{subfigure}\hfill
  \begin{subfigure}[t]{0.24\textwidth}\centering
    \includegraphics[width=\linewidth]{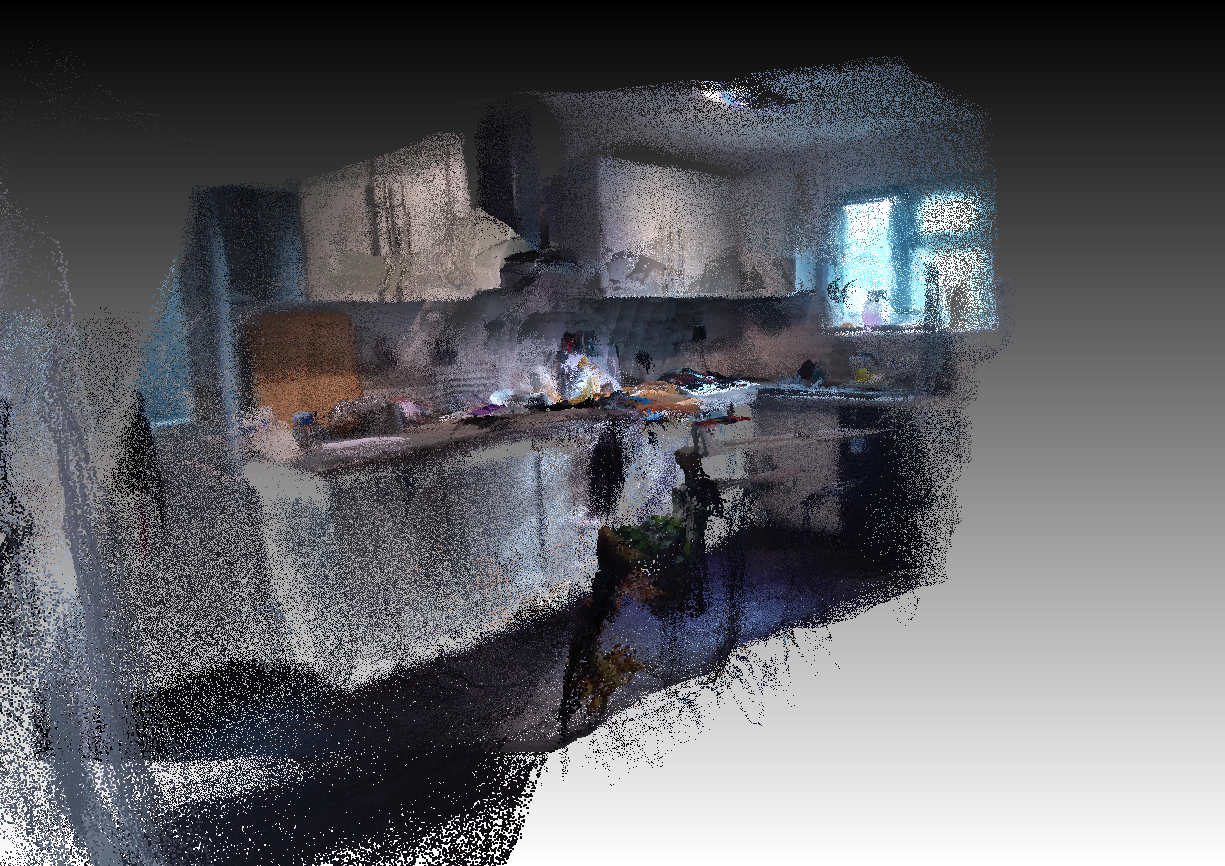}
  \end{subfigure}\hfill
  \begin{subfigure}[t]{0.24\textwidth}\centering
    \includegraphics[width=\linewidth]{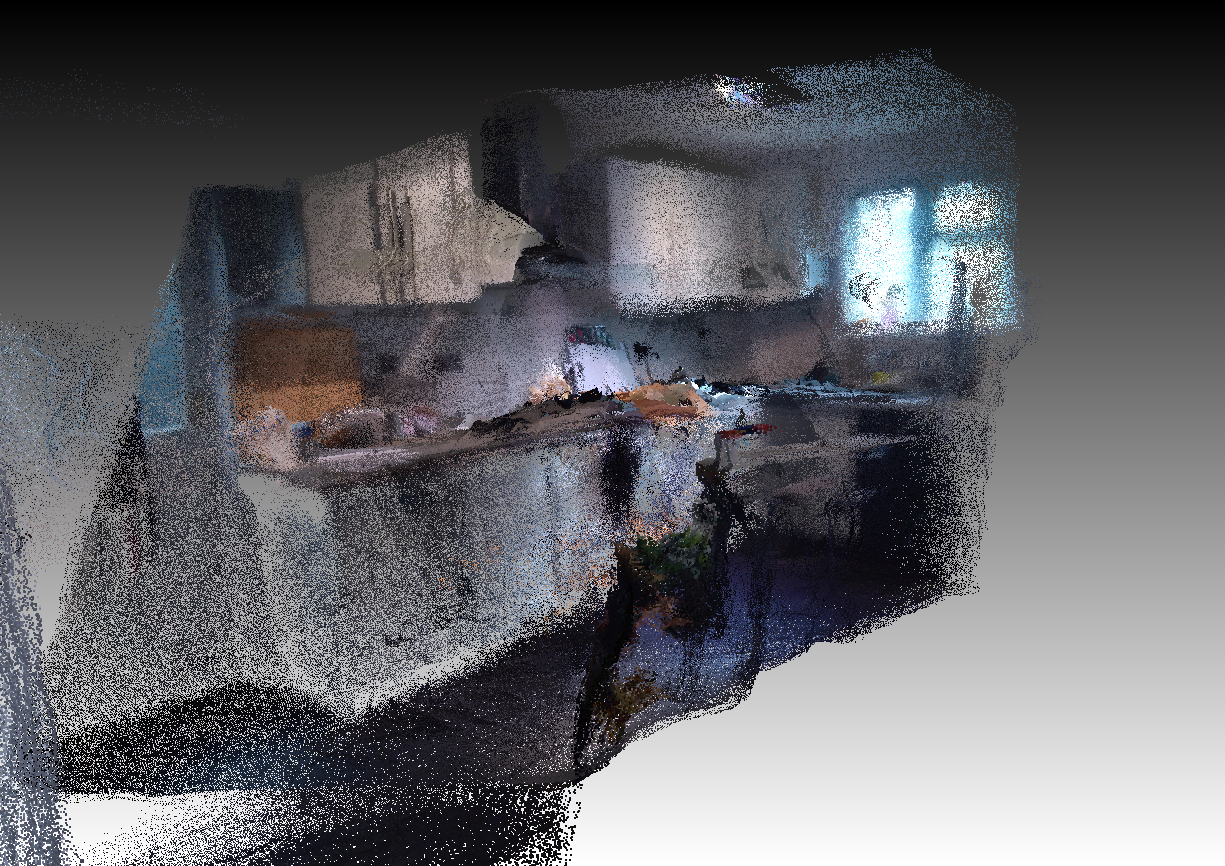}
  \end{subfigure}

  \vspace{0.8ex}

  \begin{subfigure}[t]{0.24\textwidth}\centering
    \includegraphics[width=\linewidth]{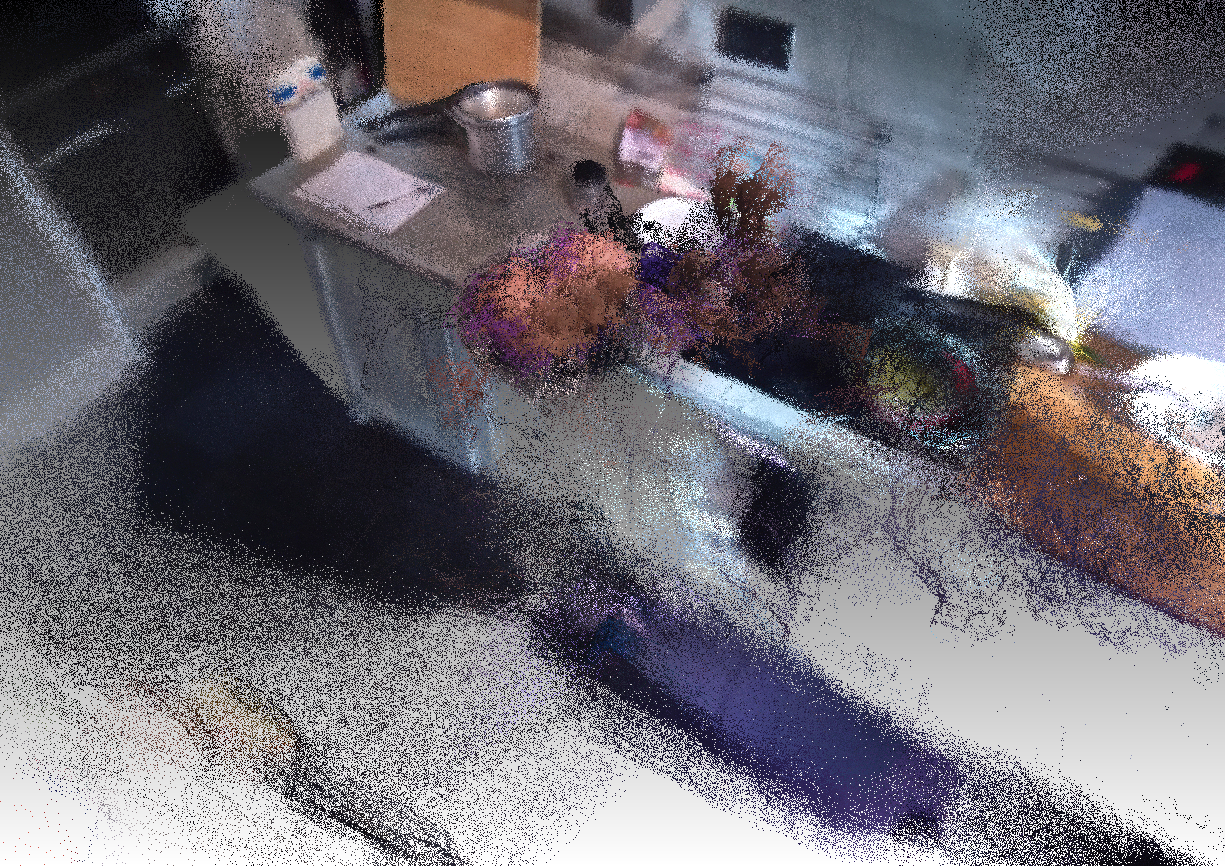}
  \end{subfigure}\hfill
  \begin{subfigure}[t]{0.24\textwidth}\centering
    \includegraphics[width=\linewidth]{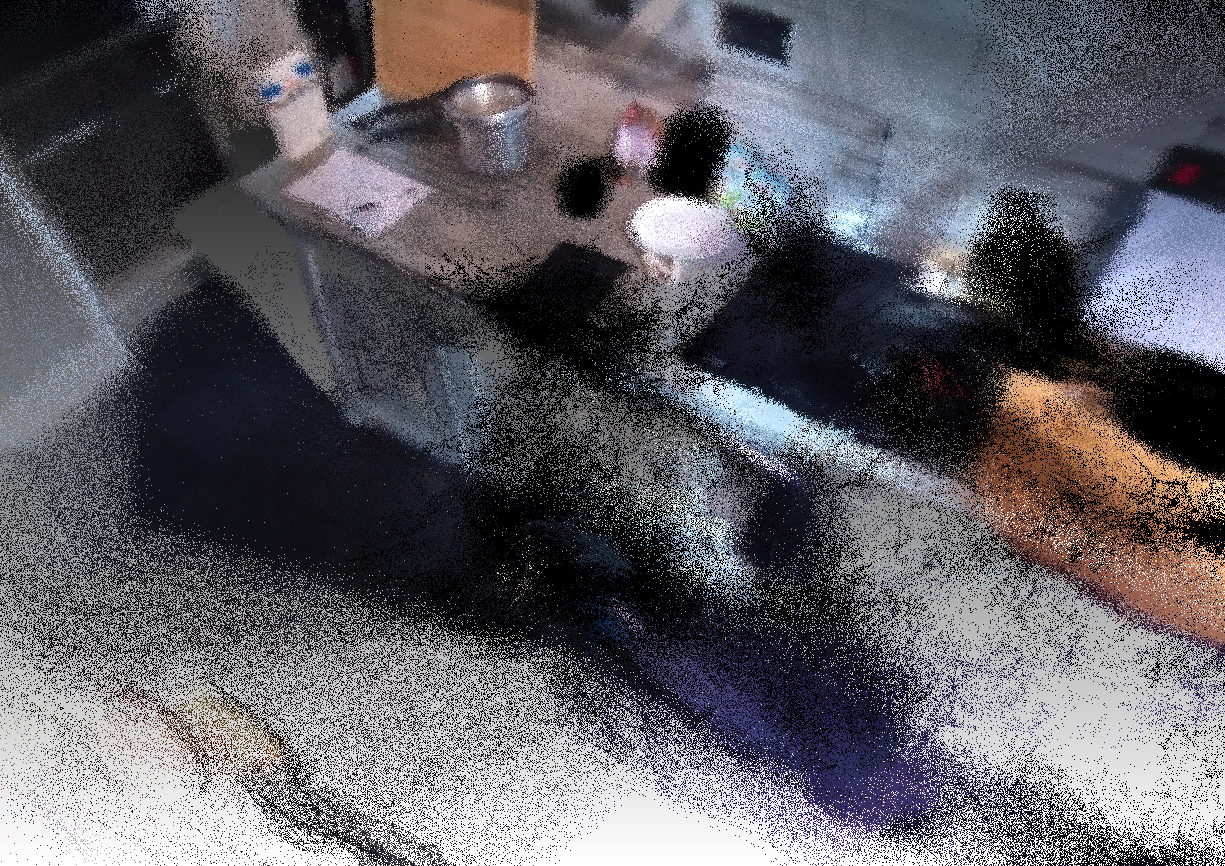}
  \end{subfigure}\hfill
  \begin{subfigure}[t]{0.24\textwidth}\centering
    \includegraphics[width=\linewidth]{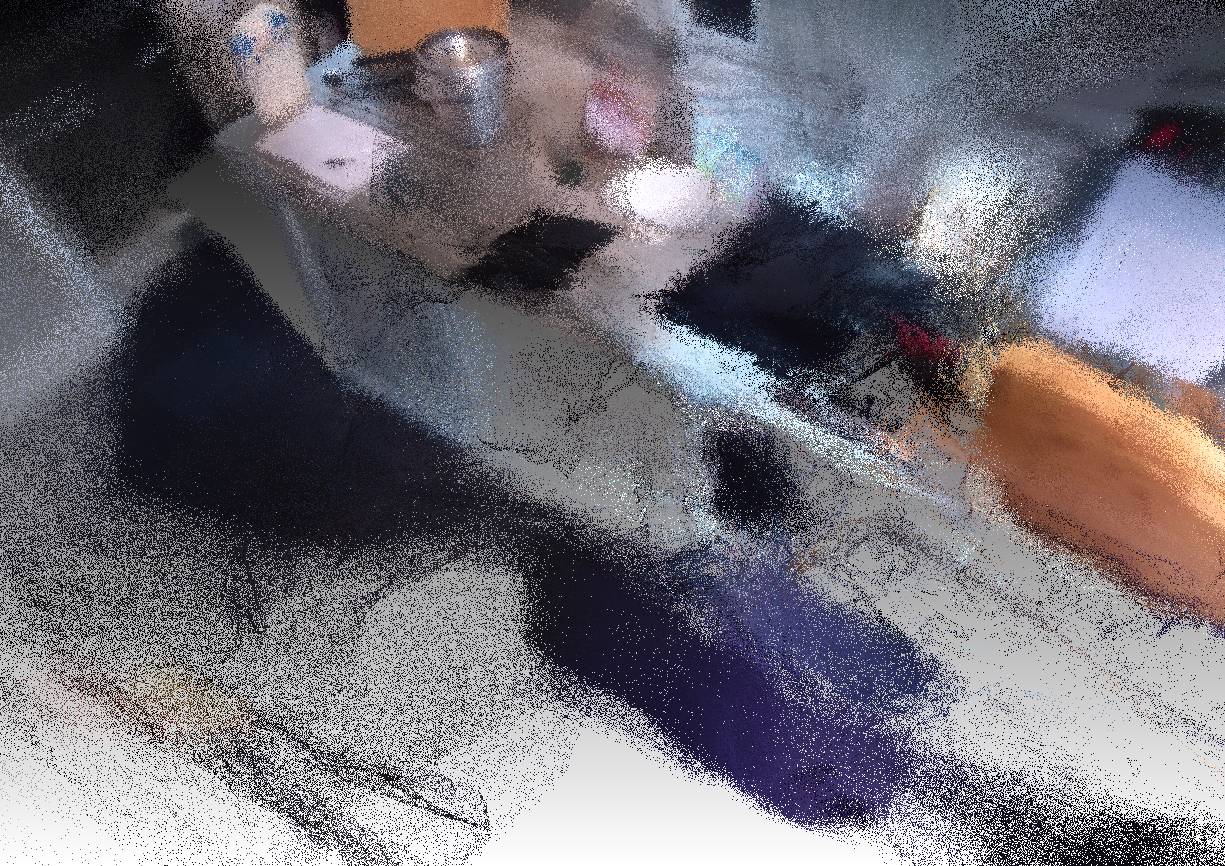}
  \end{subfigure}\hfill
  \begin{subfigure}[t]{0.24\textwidth}\centering
    \includegraphics[width=\linewidth]{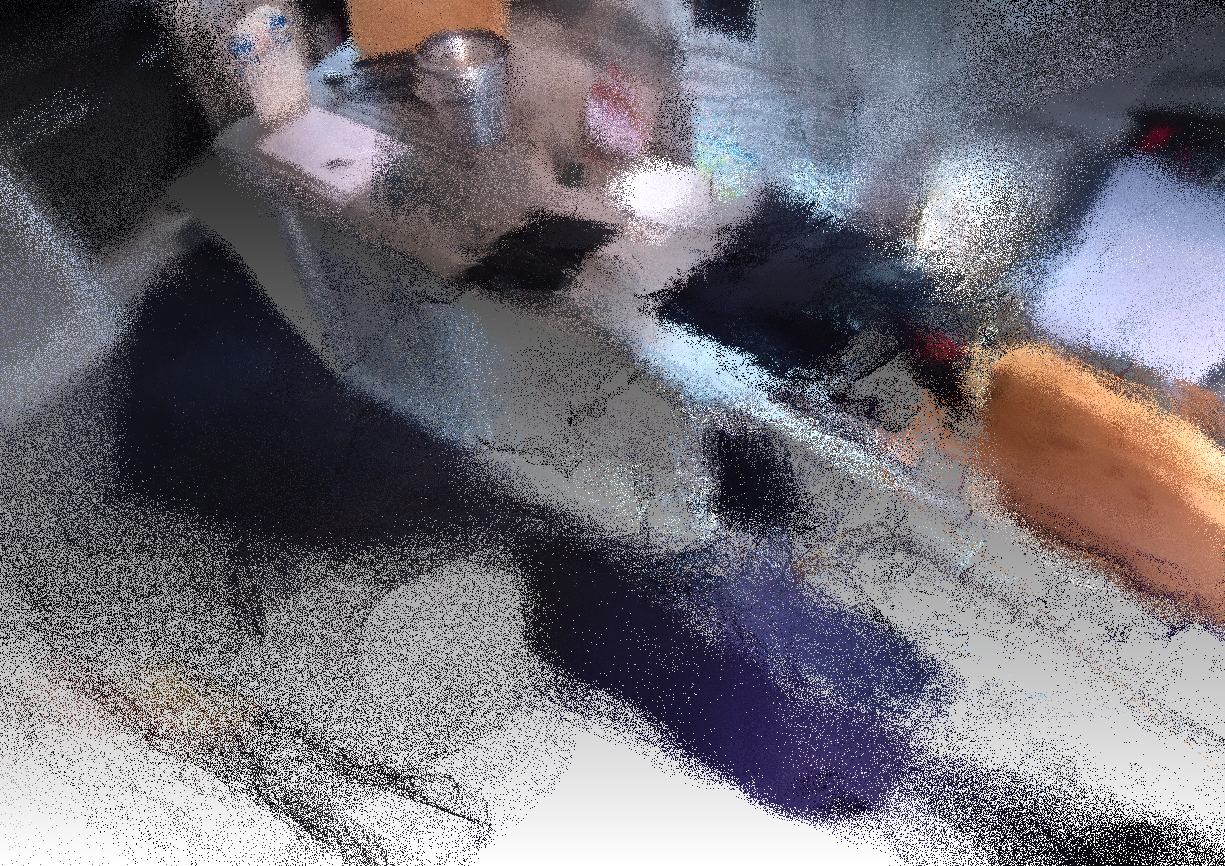}
  \end{subfigure}

  \caption{
  EPIC-KITCHENS Sense point-cloud reconstructions comparing four masking strategies across three scenes (rows).
  \emph{Baseline} (no masking) exhibits severe hand ghosting.
  \emph{Blackout} removes dynamic regions at the image level, resulting in large black voids.
  \emph{Dynamin} (ours) suppresses most ghosting while preserving sufficient static context to maintain spatial consistency.
  \emph{Overall-mask} removes all moving objects from the beginning, often leading to unstable or incomplete structure due to missing spatial cues.
  }
  \label{fig:ek_sense_all_panels}
\end{figure*}

\subsection{Limitations}
\label{sec:exp-limitations}

\paragraph{Suppression vs. completeness trade-off.}
While attention gating reduces dynamic contamination density, it can reduce depth completeness and overlap stability,
especially in interaction-heavy segments where large image areas become masked. This suggests future work on
mask-aware regularization or static-only correspondence mechanisms to preserve geometric completeness.

\paragraph{Interaction onset and object selection.}
Errors in skill segmentation, predicate selection, or object localization can propagate to mask activation.
Fully automatic performance depends on the upstream SymSkill/LLM/VLM components.

\paragraph{Mask quality and over/under-suppression.}
Imperfect SAM3 masks can leak dynamic content (residual contamination) or over-mask static pixels (missing geometry).
Improving robustness to mask noise and handling object re-entry remain future work.

\paragraph{No loop closure / global optimization.}
Our stitching is sequential overlap-based alignment without global pose-graph optimization or loop closure,
so long-range drift may remain in extremely long sequences.

{
    \small
    \bibliographystyle{ieeenat_fullname}
    \bibliography{main}
}

\clearpage
\setcounter{page}{1}

\section{Supplementary Material}
\label{sec:supp}

\begin{figure*}[t]
    \centering
    \begin{subfigure}[b]{0.48\textwidth}
        \centering
        \includegraphics[width=\textwidth]{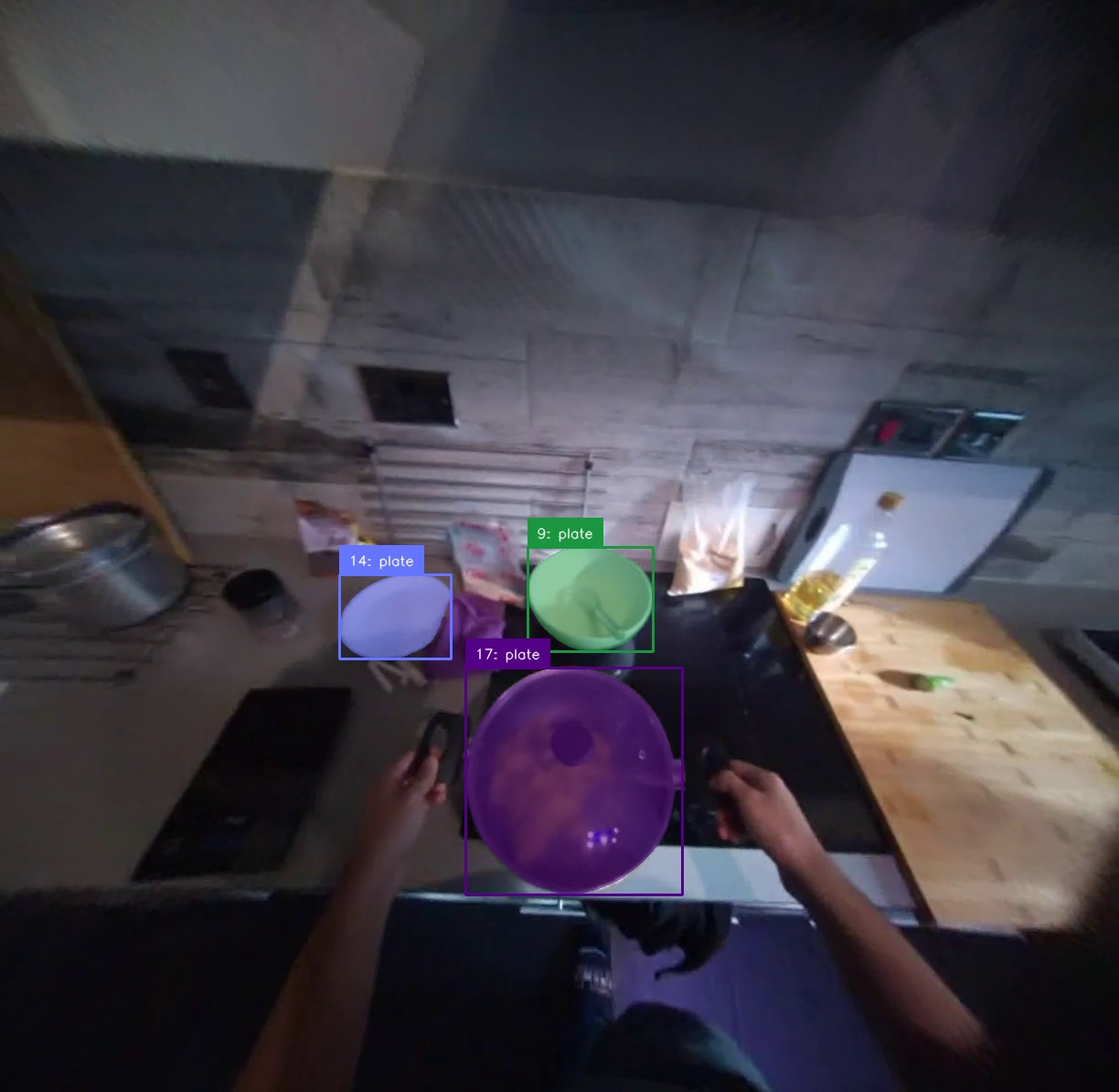}
        \caption{Baseline Grounded-SAM2 (before refinement).}
        \label{fig:supp-gsam2-before}
    \end{subfigure}
    \hfill
    \begin{subfigure}[b]{0.48\textwidth}
        \centering
        \includegraphics[width=\textwidth]{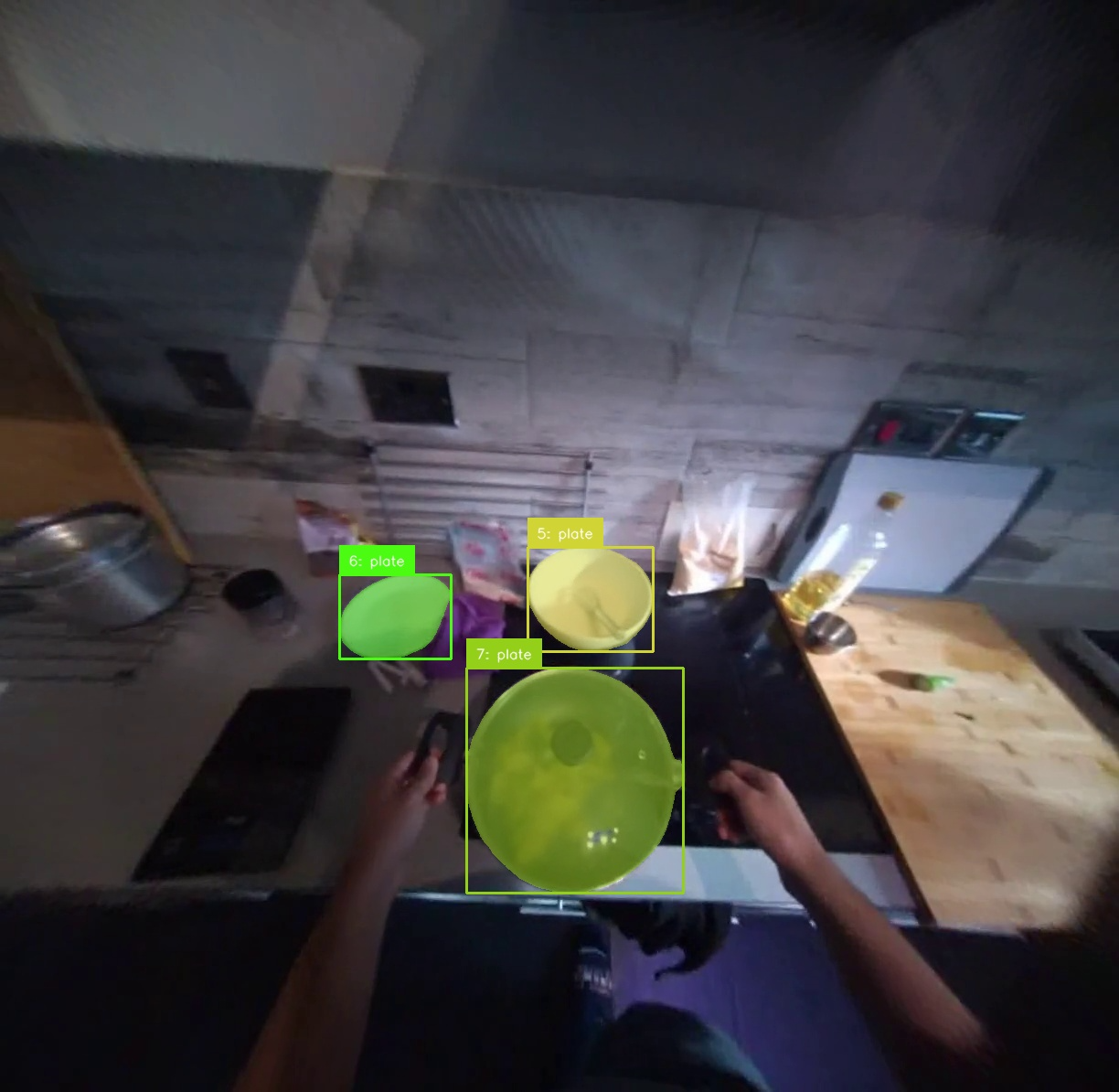}
        \caption{With post-hoc matching refinement (after).}
        \label{fig:supp-gsam2-after}
    \end{subfigure}
    \caption{\textbf{Semantic grounding vs.\ ID fragmentation (Grounded-SAM2).}
    Grounded-SAM2 provides semantic relevance but exhibits severe ID
    fragmentation in long egocentric videos; our post-hoc matching reduces the
    number of fragmented IDs but does not fully resolve failures under heavy
    occlusion and re-entry.}
    \label{fig:supp-gsam2}
\end{figure*}

\subsection{Additional Details: Tracking Attempts and Failure Modes}
\label{app:tracking}

This appendix summarizes tracking approaches explored prior to our final
interaction-aware dynamic prior, and documents their failure modes on long
egocentric videos. These details support \S\ref{sec:exp-failures}.

\paragraph{Grounded-SAM2: semantic grounding with ID fragmentation.}
Grounded-SAM2 combines text-conditioned grounding (e.g., Grounding-DINO) with
SAM-style mask prediction, enabling semantically targeted segmentation. In long
videos with occlusions and viewpoint changes, we observed frequent ID switches
and fragmentation: the same physical object could be assigned many short-lived
IDs. To partially mitigate this, we implemented a post-hoc matching module that
re-associates detections to tracks using multiple cues:
\begin{equation}
S \;=\; \alpha\cdot\mathrm{IoU} \;+\; \beta\cdot\exp(-d_{\mathrm{geo}}) \;+\; \gamma\cdot \mathrm{sim}_{\mathrm{DINO}},
\end{equation}
where $\mathrm{IoU}$ measures mask overlap with recent frames, $d_{\mathrm{geo}}$
captures geometric consistency (e.g., centroid and box changes), and
$\mathrm{sim}_{\mathrm{DINO}}$ is cosine similarity of DINOv2 features computed
over the masked region. While this reduced fragmentation (e.g., max IDs
$17\rightarrow 7$ on a representative kitchen clip), track breaks persisted
under long occlusions and re-entry, limiting downstream use for reconstruction.

\paragraph{SAM-Track + DeAOT: temporal stability with over-segmentation.}
We also tested SAM-Track, which initializes masks on keyframes and uses DeAOT for
temporal propagation with long-term memory. This improves ID retention and can
recover objects after temporary occlusions. However, because SAM-Track lacks
semantic filtering, it progressively segments and tracks many irrelevant regions
in long videos, accumulating hundreds of tracks and consuming substantial memory
and compute. This made outputs difficult to curate and impractical as a robust
dynamic prior for mapping.

\paragraph{Takeaway for the final pipeline.}
These experiments suggest a semantic-vs-temporal trade-off: methods with strong
semantic grounding fragment IDs under occlusion, while methods with strong
temporal propagation tend to over-segment without category awareness. For our
primary goal (static scene reconstruction), we therefore use SAM3 primarily as a
mask generator and tracker to identify dynamic regions, combined with
interaction-aware activation to prevent moved objects from contaminating the
static map.

\begin{figure}[t]
    \centering
    \includegraphics[width=\linewidth]{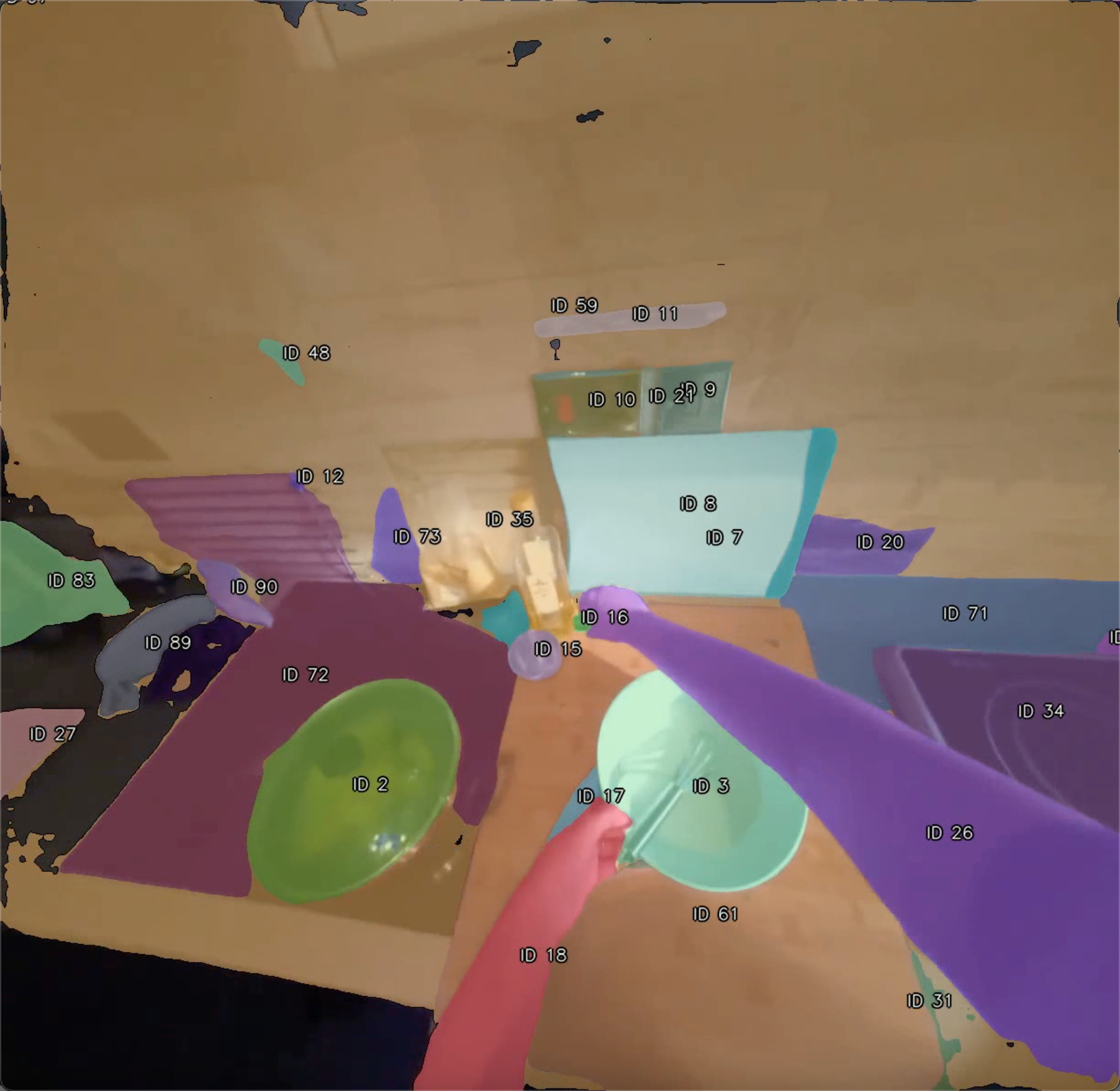}
    \caption{\textbf{Temporal stability vs.\ over-segmentation (SAM-Track + DeAOT).}
    While long-term memory improves ID retention, the method progressively
    segments and tracks many irrelevant regions in long videos, yielding an
    explosion in tracked objects without semantic filtering.}
    \label{fig:supp-samtrack}
\end{figure}

\subsection{Exploratory Results: SAM3D for Single-Frame Object Reconstruction}
\label{app:sam3d}

Although our main scope is static scene reconstruction, we explored SAM3D as a
future direction for object-level reconstruction. SAM3D can generate textured 3D
meshes from a single RGB image, which reduces reliance on long-horizon tracking.
On representative frames from kitchen scenes (plates, spatulas, spray bottles)
and ScanNet++ objects, SAM3D produced plausible geometry and textures. A natural
future direction is to combine SAM3 tracking for temporal identification and
keyframe selection with SAM3D for per-frame mesh reconstruction, followed by
mesh registration and fusion.

\begin{figure*}[t]
    \centering
    \begin{subfigure}[b]{0.24\textwidth}
        \centering
        \includegraphics[width=\textwidth]{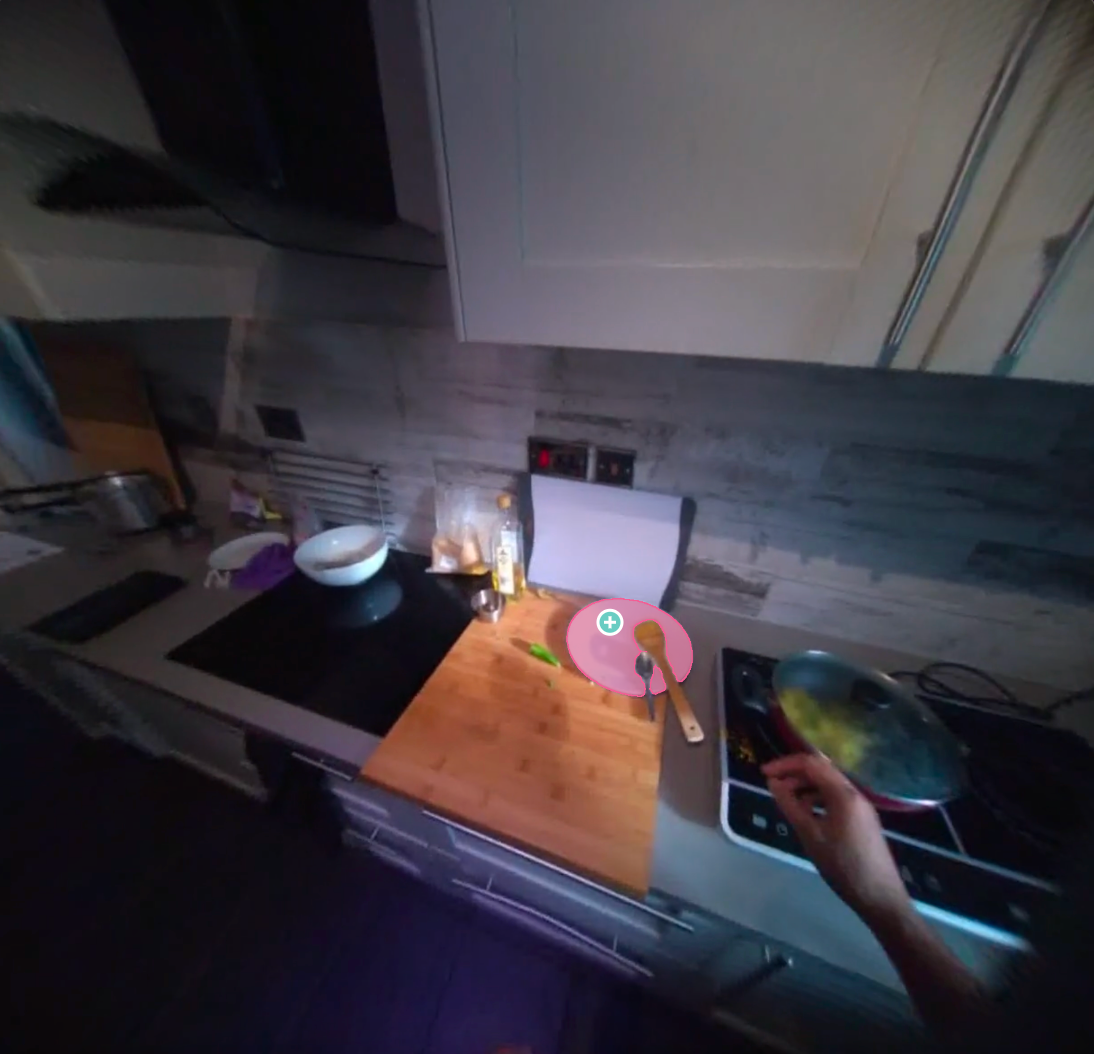}
        \caption{Plate (input)}
    \end{subfigure}
    \hfill
    \begin{subfigure}[b]{0.24\textwidth}
        \centering
        \includegraphics[width=\textwidth]{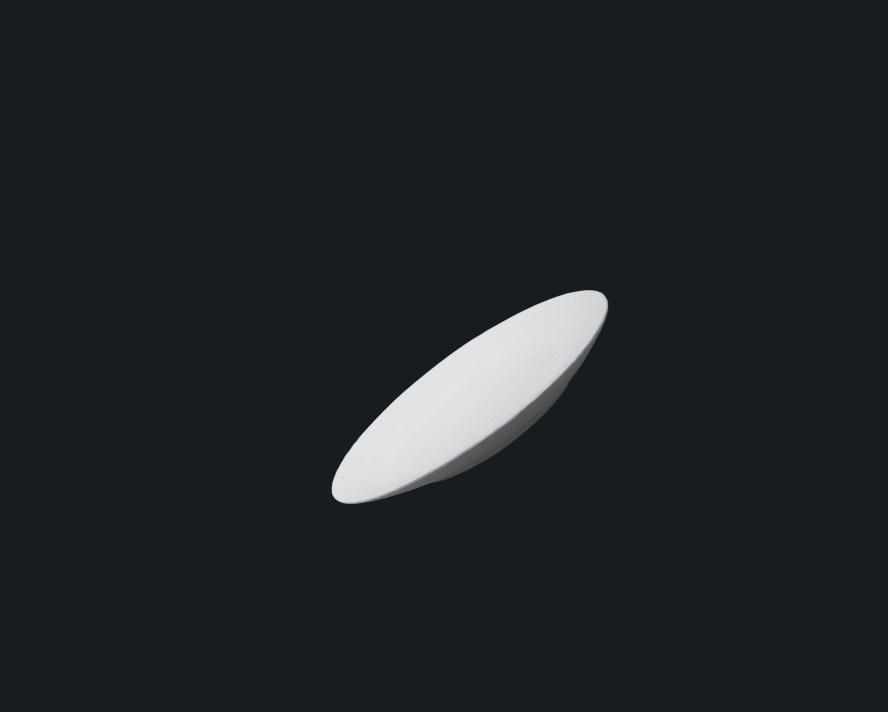}
        \caption{Plate (mesh)}
    \end{subfigure}
    \hfill
    \begin{subfigure}[b]{0.24\textwidth}
        \centering
        \includegraphics[width=\textwidth]{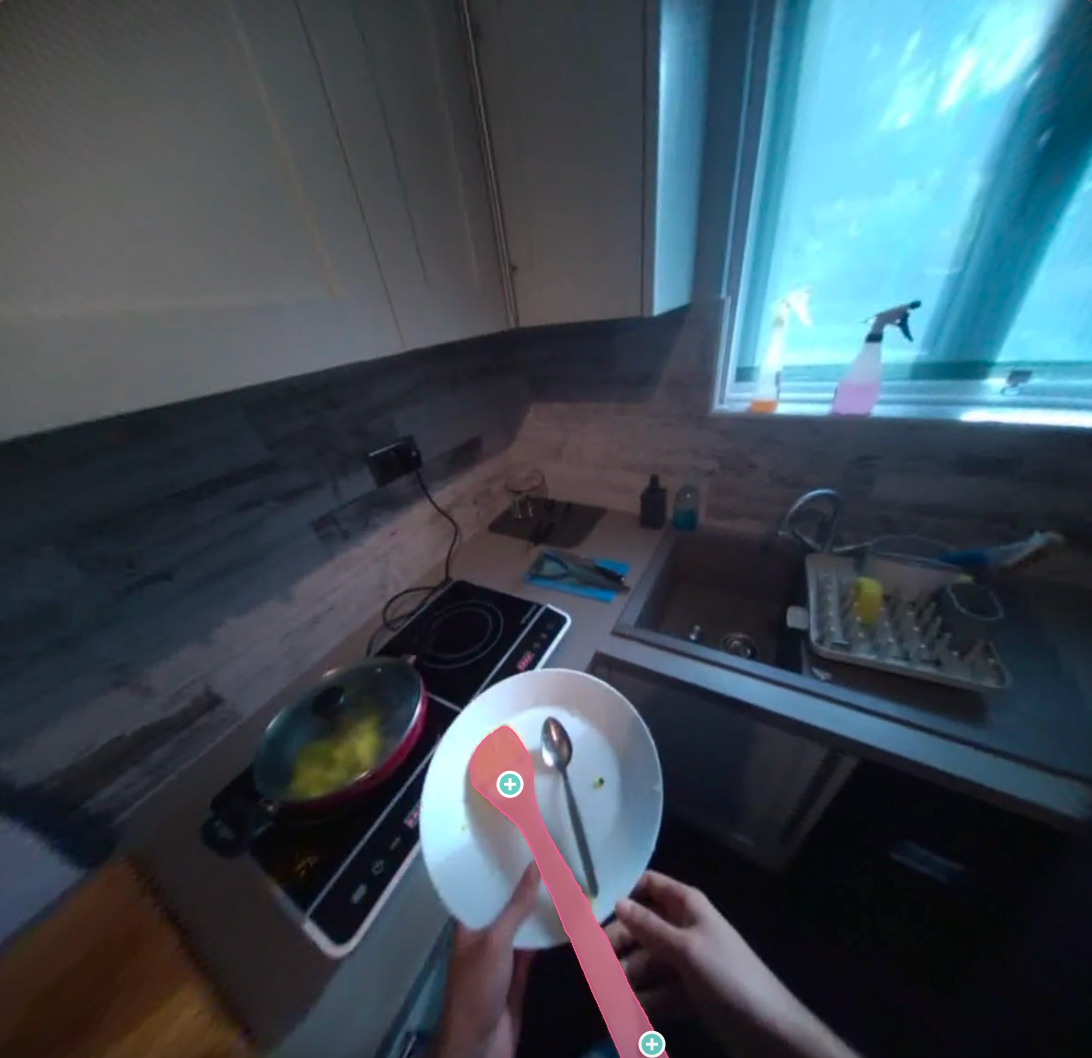}
        \caption{Spatula (input)}
    \end{subfigure}
    \hfill
    \begin{subfigure}[b]{0.24\textwidth}
        \centering
        \includegraphics[width=\textwidth]{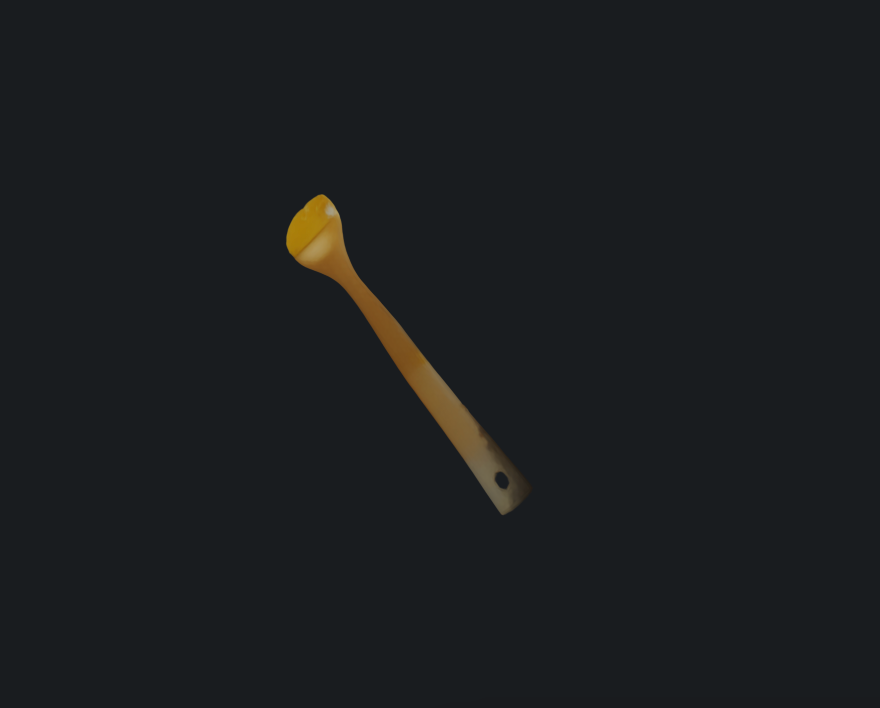}
        \caption{Spatula (mesh)}
    \end{subfigure}
    
    \vspace{0.25cm}
    
    \begin{subfigure}[b]{0.24\textwidth}
        \centering
        \includegraphics[width=\textwidth]{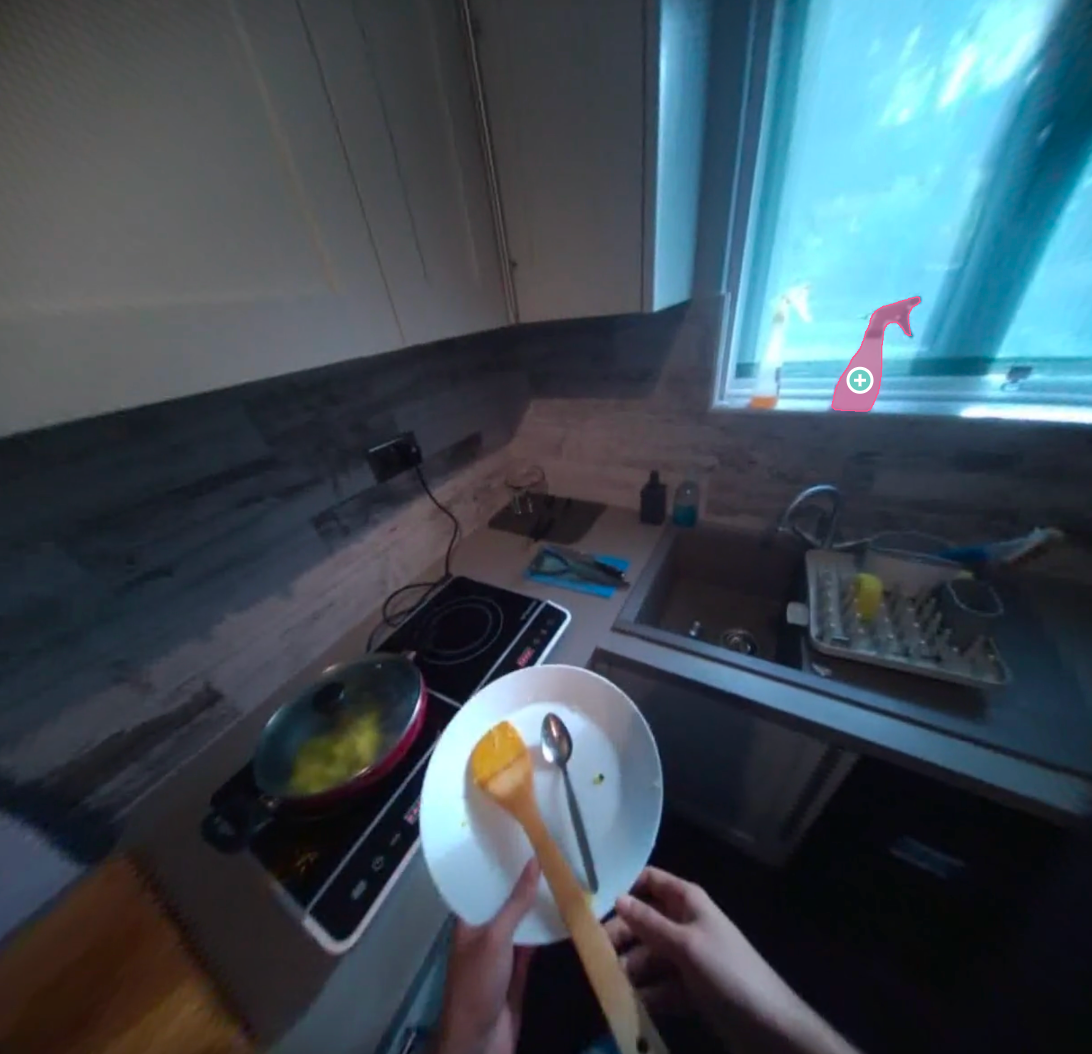}
        \caption{Spray bottle (input)}
    \end{subfigure}
    \hfill
    \begin{subfigure}[b]{0.24\textwidth}
        \centering
        \includegraphics[width=\textwidth]{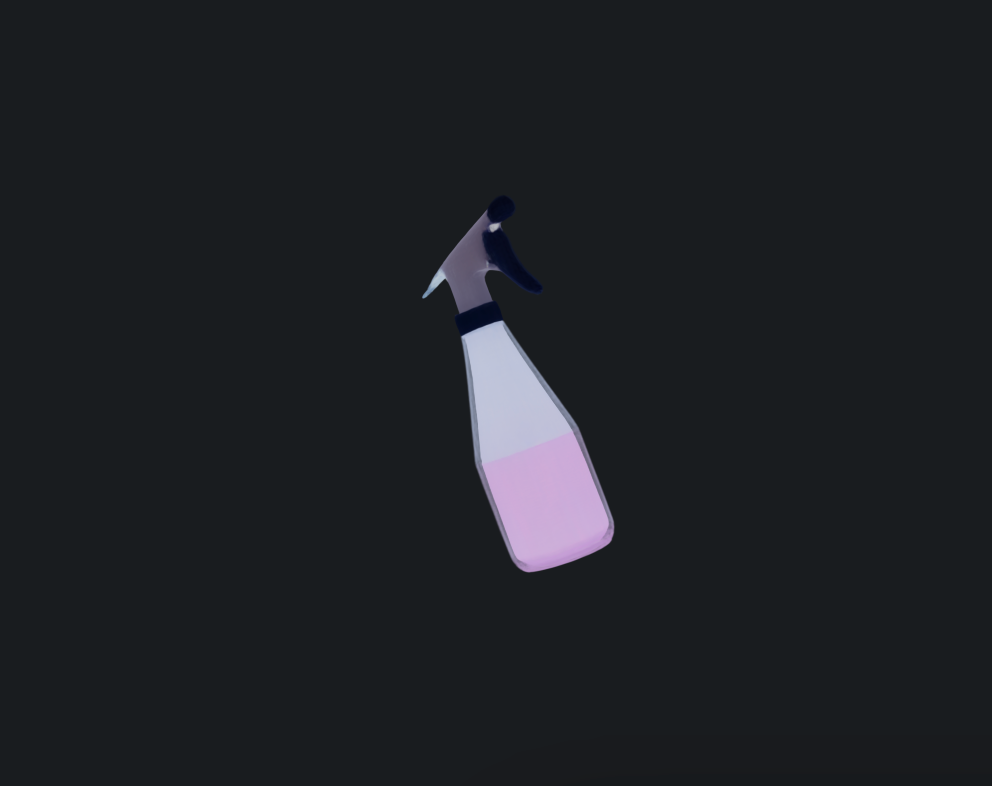}
        \caption{Spray bottle (mesh)}
    \end{subfigure}
    \hfill
    \begin{subfigure}[b]{0.24\textwidth}
        \centering
        \includegraphics[width=\textwidth]{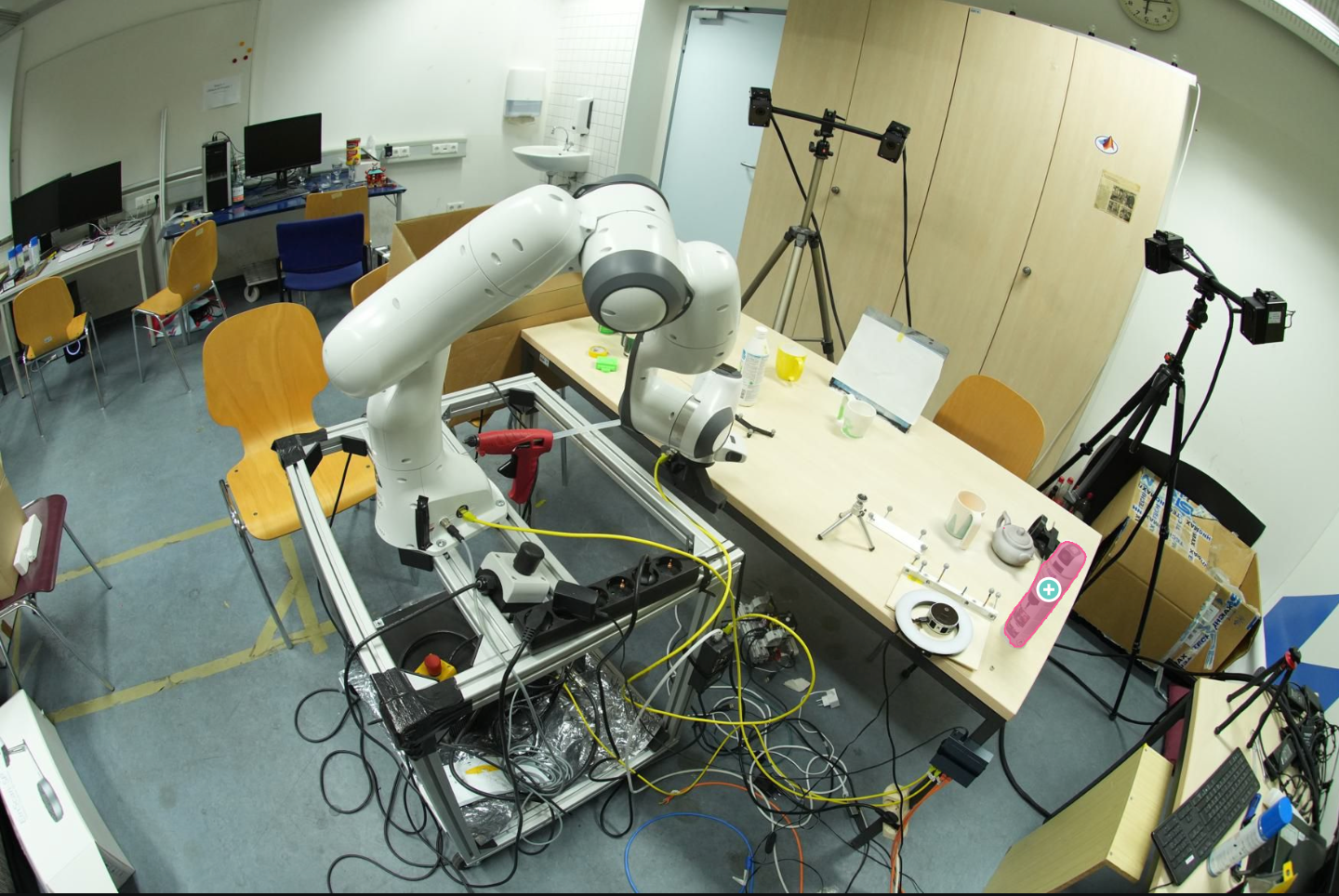}
        \caption{Bottle (input)}
    \end{subfigure}
    \hfill
    \begin{subfigure}[b]{0.24\textwidth}
        \centering
        \includegraphics[width=\textwidth]{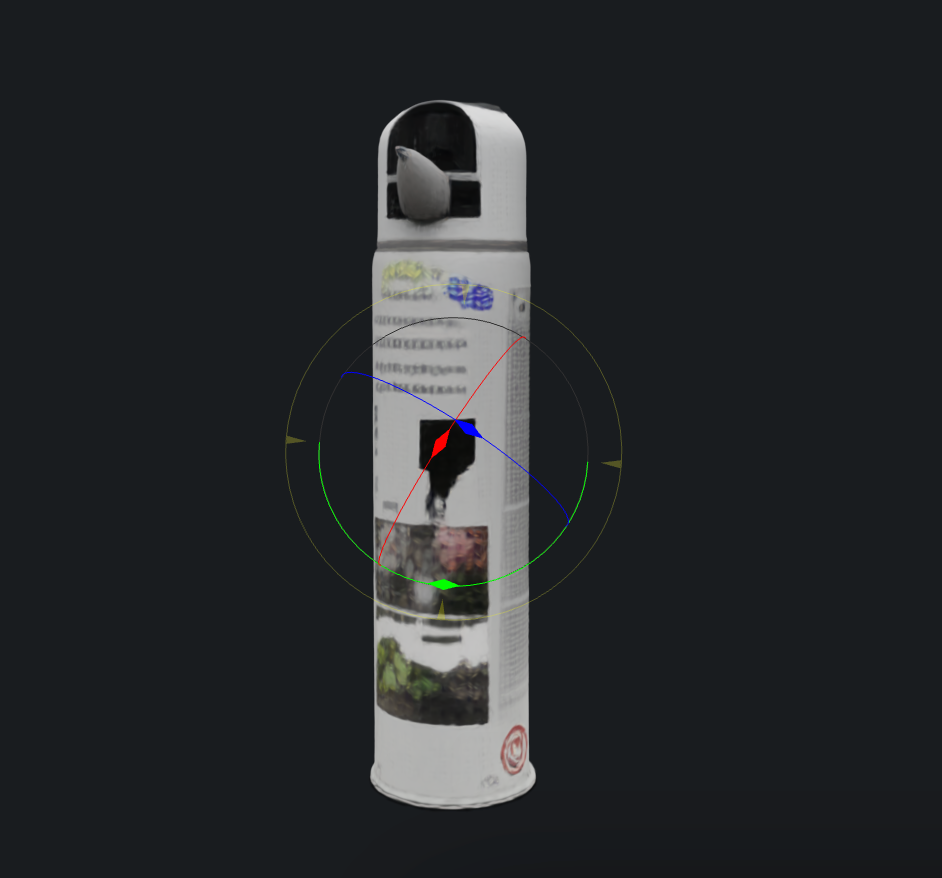}
        \caption{Bottle (mesh)}
    \end{subfigure}
    
    \caption{\textbf{SAM3D single-frame reconstruction (exploratory).} For each
    object category, we show the input image (with mask) and the generated
    textured mesh. While outside our main scope, these results suggest a future
    direction for object-level reconstruction on top of our static mapping
    pipeline.}
    \label{fig:supp-sam3d}
\end{figure*}

\end{document}